\documentclass[preprint,12pt]{elsarticle}
\usepackage{algorithm}
\usepackage{algpseudocode}  
\usepackage{graphicx}
\usepackage{subcaption}
\usepackage{comment} 
\usepackage{makecell} 
\usepackage{xcolor}

\usepackage{amssymb}
\usepackage{amsmath}

\journal{Transportation Research Part C}

\begin{document}

\begin{frontmatter}

\title{Extended Visibility of Autonomous Vehicles via Optimized Cooperative Perception under Imperfect Communication}

\author[cu]{Ahmad Sarlak}
\author[mitll]{Rahul Amin}
\author[cu]{Abolfazl Razi\corref{cor1}}

\cortext[cor1]{Corresponding author: Abolfazl Razi, Email: arazi@clemson.edu}

\cortext[email]{Email addresses: asarlak@g.clemson.edu (Ahmad Sarlak), rahul.amin@ll.mit.edu (Rahul Amin), arazi@clemson.edu (Abolfazl Razi). This material is based upon the work supported by the National Science Foundation under Grant Numbers 2008784 and 2204721.}

\affiliation[cu]{organization={School of Computing, Clemson University},
                 city={Clemson},
                 state={SC},
                 country={USA}}

\affiliation[mitll]{organization={Lincoln Laboratory, Massachusetts Institute of Technology},
                 city={Lexington},
                 state={MA},
                 country={USA}}

\begin{abstract}
Autonomous Vehicles (AVs) rely on individual perception systems to navigate safely. However, these systems face significant challenges in adverse weather conditions, complex road geometries, and dense traffic scenarios. Cooperative Perception (CP) has emerged as a promising approach to extending the perception quality of AVs by jointly processing shared camera feeds and sensor readings across multiple vehicles. This work presents a novel CP framework designed to optimize vehicle selection and networking resource utilization under imperfect communications. Our optimized CP formation considers critical factors such as the helper vehicles' spatial position, visual range, motion blur, and available communication budgets. 
Furthermore, our resource optimization module allocates communication channels while adjusting power levels to maximize data flow efficiency between the ego and helper vehicles, considering realistic models of modern vehicular communication systems, such as LTE and 5G NR-V2X. We validate our approach through extensive experiments on pedestrian detection in challenging scenarios, using synthetic data generated by the CARLA simulator. The results demonstrate that our method significantly improves upon the perception quality of individual AVs with about 10\% gain in detection accuracy. This substantial gain uncovers the unleashed potential of CP to enhance AV safety and performance in complex situations.
\end{abstract}

\begin{graphicalabstract}
\includegraphics[width=\textwidth]{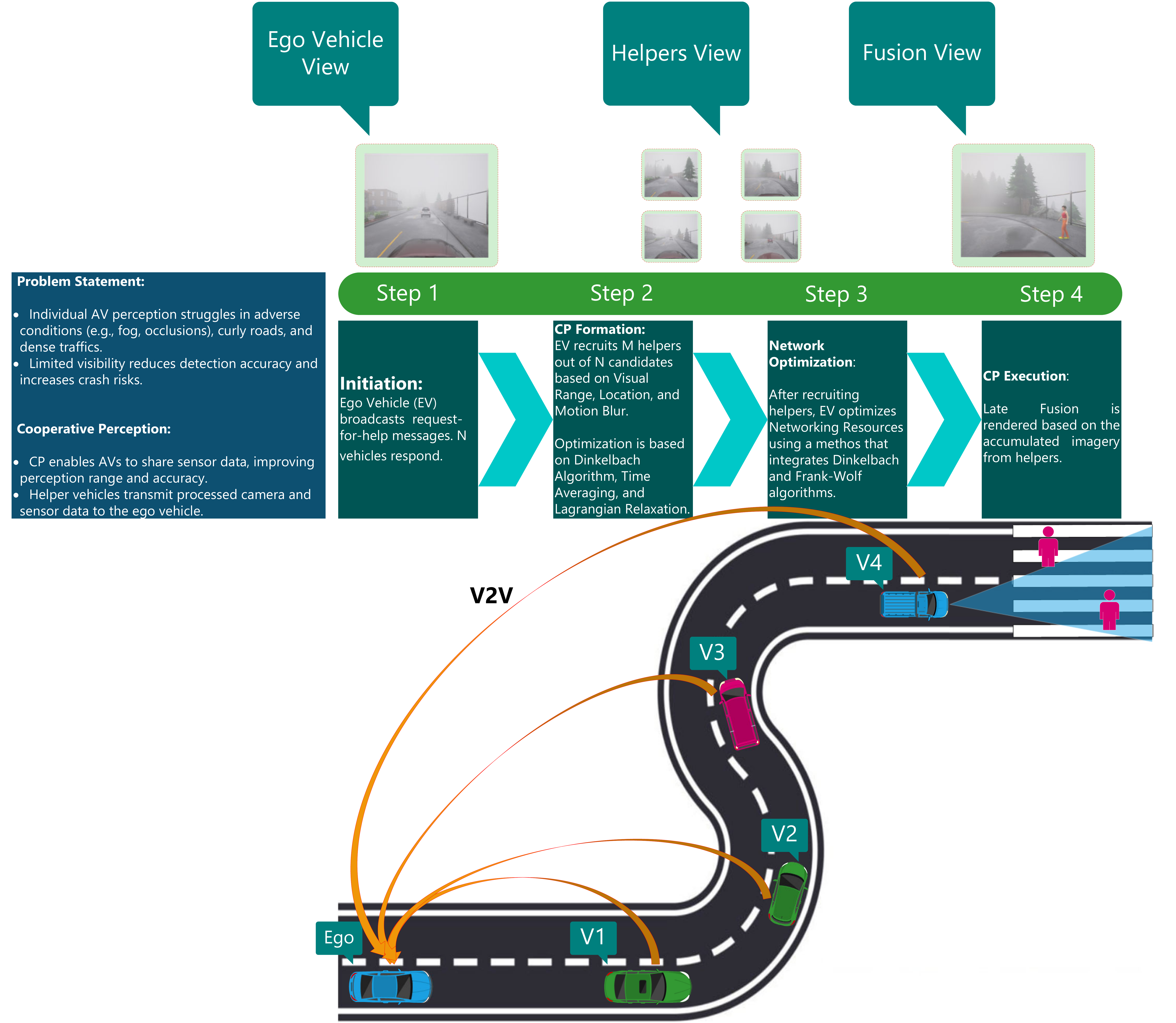}
\end{graphicalabstract}

\begin{highlights}
\item Enhance perception quality of autonomous vehicles via cooperative perception.

\item Optimize cooperative perception quality and situational awareness via informed helper selection based on their visual range, relative position, motion blur, and channel quality under imperfect communication.

\item  Regulate networking resource utilization and power levels for improved V2V communication and perception quality.


\item Introduce a novel lightweight optimization framework by integrating Dinkelbach's algorithm and Frank–Wolfe's algorithm for optimized networking appropriate for real-time vehicular communication.

\item Demonstrate significant gain with simulated scenarios for pedestrian detection, as an exemplary application.
\end{highlights}

\end{frontmatter}

\section{Introduction}

CP arises as a promising paradigm in AVs, addressing the inherent limitations of individual vehicle perception, especially in adverse weather conditions, complex traffic scenarios, and environments with limited visibility \cite{gueriau2016assess, sarlak2023cooperative, chen2022milestones}. CP involves sharing camera feeds and sensor readings of multiple AVs within a network, enabling enhanced situational awareness through the collective processing of shared perception inputs \cite{li2024cooperative, yang2021cooperative}. 
The effectiveness of CP lies in its ability to address the inherent challenges faced by individual AV perception systems in scenarios involving low illumination, occlusions, adverse weather conditions (e.g., fog, and heavy rain), and complex road geometries (e.g., winding roads). In such situations, relying solely on a single vehicle’s sensors and visual feeds often results in perception errors, compromising navigation and safety. By enabling vehicles to share and collaboratively process information, CP provides a robust pathway to overcome these limitations, significantly enhancing safety, resilience, and decision-making for AVs.

The concept of enhanced perception quality via CP is shown in multiple prior works, \cite{sarlak2024enhanced, li2023learning, lin2024v2vformer}. Nonetheless, there exists no substantial work, to the best of our knowledge, that addresses the critical process of identifying and recruiting helper vehicles under given conditions, so that the ultimate perception quality is maximized. This paper aims to formalize and solve the process of selecting a subset of volunteer helper vehicles that can offer the highest contributions to the ultimate quality of the CP systems when joining the process, taking into account their spatial arrangements and locations (e.g., distance to the front vehicle), relative velocities, channel quality, and available networking resources.

\begin{figure}[H]
\centering
\begin{subfigure}{.8\columnwidth}
  \includegraphics[width=\textwidth]{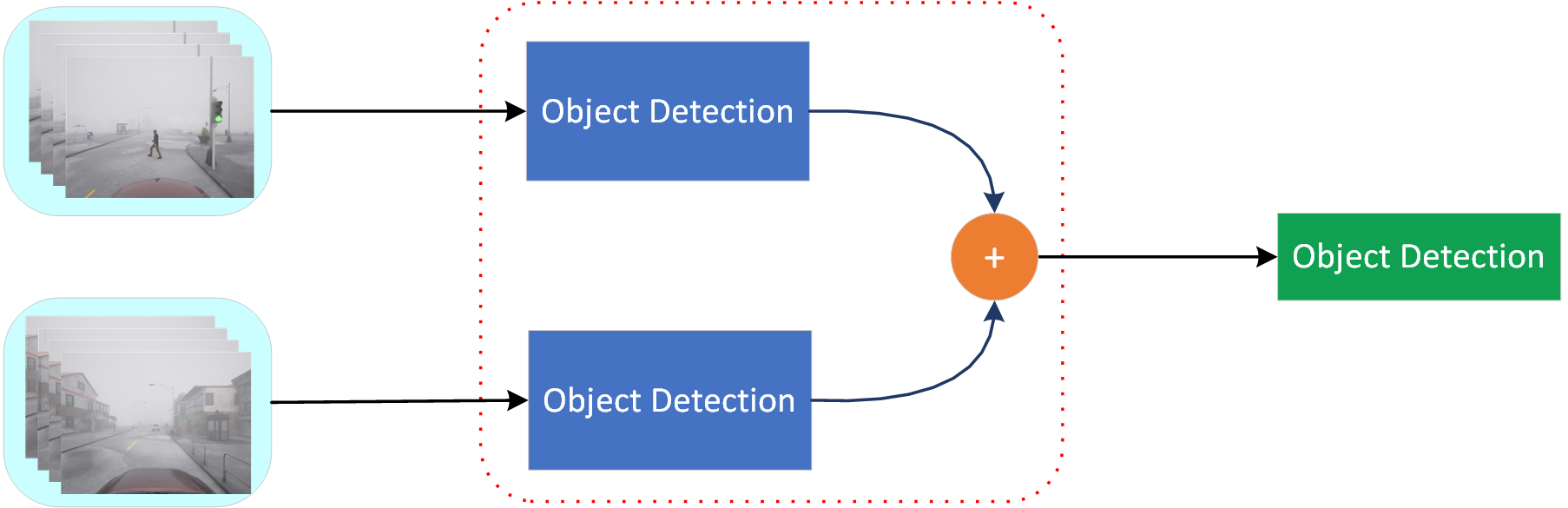}
  \caption{}
\end{subfigure}

\begin{subfigure}{.8\columnwidth}
  \includegraphics[width=\textwidth]{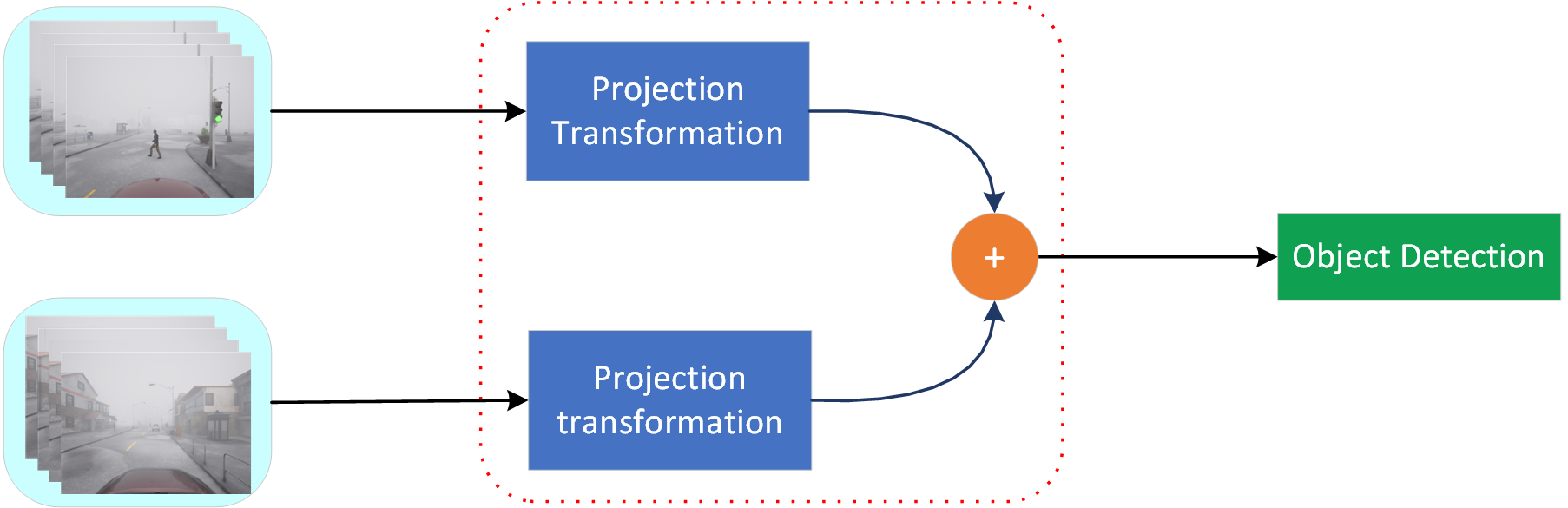}
  \caption{}
\end{subfigure}

\begin{subfigure}{.8\columnwidth}
  \includegraphics[width=\textwidth]{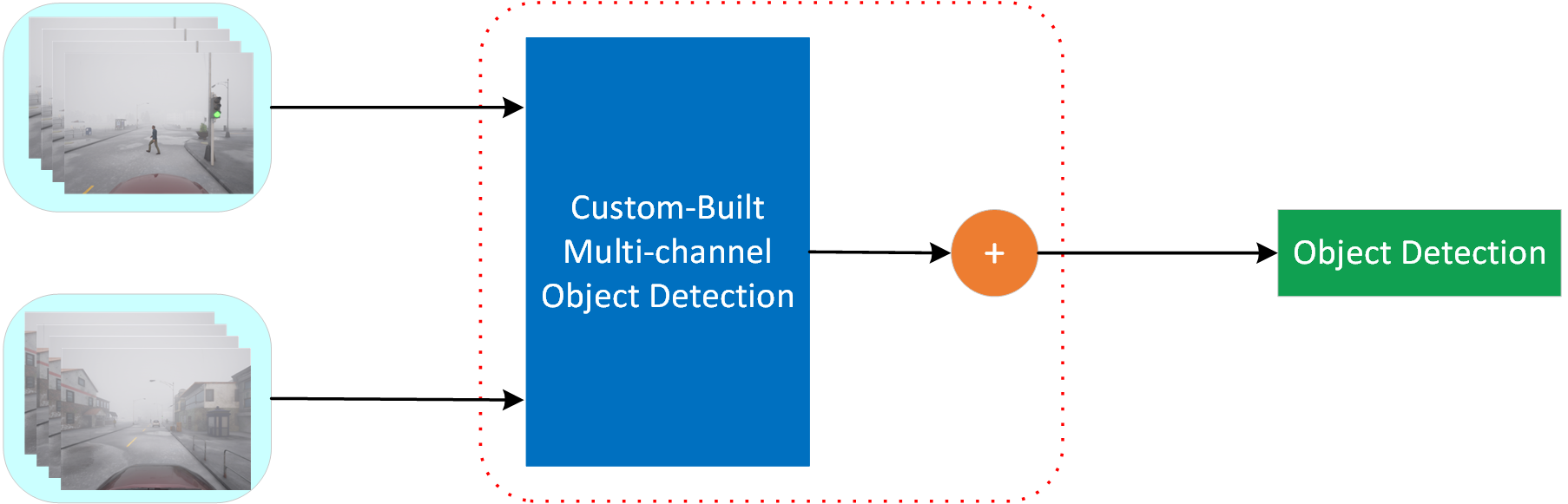}
  \caption{}
  \end{subfigure}
\vspace{-0.5cm}
\caption{Three fusion approaches used in CP, including (a) early fusion, (b) late fusion, and (c) integrative analysis.}
\label{fig:fusion}
\vspace{-0.2 in}
\end{figure}

A typical CP framework employs Vehicle-to-Vehicle(V2V) communication technology to share sensory data among vehicles to enable collective processing \cite{chang2023using, xu2022opv2v, xu2022cobevt, sarlak2023diversity}. There exist three mainstream approaches to fusing the collected shared visual information, as illustrated in Figure \ref{fig:fusion}. The first approach is \textit{early fusion}, where the raw images captured by multiple vehicles are translated to the same field of view through optimized image projection (also called image warping and image transformation) to offer the same point of view. These completely aligned images are then merged into a single high-quality image to be used by the downstream learning-based tasks such as Object Detection (OD), depth estimation, reinforcement learning (RL), etc. (Fig. \ref{fig:fusion} (a)). This can be viewed as mixing information at the pixel level. The main issue of this approach is the sensitivity of the fused images to the projection parameters. 
The second approach is \textit{late fusion}, where each image (or video feed) is processed individually predominantly by the learning-based module. Then the obtained results are combined to achieve more reliable results (Fig. \ref{fig:fusion} (b)). For instance, one may execute OD on individual images to obtain object classes and bounding boxes to be fused to get more precise results. The key advantage of these two methods is their straightforward architecture that enables easy integration with existing deep learning (DL)-based visual processing methods \cite{zhang2021distributed}. The third approach is integrative processing, which entails developing new DL \textit{architectures} capable of jointly processing multiple image/video channels (Fig. \ref{fig:fusion} (c)). Such methods can typically yield better results for specific tasks if elegantly designed. The downside of the approach is the need for more sophisticated custom-designed DL architectures that can limit their applicability.

In this work, we adopt the second approach, late fusion, for its robust operation, reasonable performance, implementation convenience, and above all, for its seamless integration with existing DL architectures. A unique feature of our method is optimized helper selection, where we propose a framework to select an optimal set of helper vehicles, which can collectively achieve the highest perception quality. The key intuition is to select the helper vehicles (out of the volunteers) that collectively realize the longest extension for the go vehicle's visibility while accounting for other parameters such as their motion blur, and distance to the ego vehicle. This approach is distinct from existing CP methods, which often overlook the vehicle selection process. Indeed, most prior CP works strive to fuse multiple perception modalities (such as \cite{qiao2023adaptive} that integrates visual information with LiDAR point clouds, and \cite{xiang2023multi} that combines images with equal characteristics). To the best of our knowledge, no substantial prior work has explored the integration of image channels under heterogeneous conditions and perspectives for OD in the realm of autonomous driving. We have shown that the gain of CP can be significantly improved by an optimized selection of helper vehicles.

Furthermore, most existing methods have taken unrealistic assumptions, such as perfect communication and unlimited transmission resources \cite{bian2019reducing, gould2023information, dey2016vehicle}, not reflecting the constraints of real-world scenarios. Indeed, vehicular communication channels often experience higher interruption rates than other communication systems \cite{liu2023fault, ren2024interruption, du2016information}, underscoring the necessity of taking communication imperfections into account. Only a few studies have addressed imperfect communications in CP, such as \cite{xu2022v2x, hakim2023ccpav}, where the authors propose V2X-ViT, a vision transformer-based CP framework leveraging Vehicle-to-Everything (V2X) communication to enhance AV perception for OD. While the authors attempted to address V2X challenges, their sole focus was achieving robustness to communication delays, overlooking other essential communication aspects (packet drop rate, throughput, etc.). 
Likewise, \cite{thandavarayan2023scalable, li2023learning} explores the impact of imperfect V2X communication on CP for OD. The authors in \cite{li2023learning} proposed an innovative approach with an LC-aware Repair Network (LCRN) and a V2V Attention Module (V2VAM) to mitigate the adverse effects of imperfect communications. They significantly improved the detection performance demonstrated on the OPV2V dataset. Nonetheless, they only attempted to adjust their method for varying channel quality without incorporating channel quality into the vehicle selection process.

In this work, we propose a novel approach to CP that overcomes the limitations of the previous methods by enabling selective CP, in which the helper vehicles are selected based on their visual information and channel quality. The networking parameters are adopted from the NR-V2X side-link communication used for V2V communication.  
By doing so, we achieve enhanced situational awareness through higher computer vision-based detection accuracy under imperfect and constrained communication scenarios compared to computer vision methods that blindly or randomly select helper vehicles. While we validate our method using our own implementation of \textit{late fusion} for OD, the proposed helper recruit mechanism is universal and can be integrated with arbitrary CP architectures. It can also be used for other applications beyond pedestrian detection, including speed estimation, lane detection, vehicle classification, traffic sign detection, traffic light interpretation, etc.

More specifically, we select helper vehicles based on their contribution to extending the visual range of the ego vehicle, considering their geo-locations and their imperfect channel conditions. Additionally, we take into account the vehicles' relative velocities because the motion blur impacts their imaging quality \cite{xiao2021vehicle, cortes2018velocity, ye2020federated} and consequently the ultimate accuracy of the developed CP-based OD accuracy.
To simplify the problem, we take a two-step approach where we first determine the optimal set of vehicles based on factors such as the vehicles' location, visual range, motion blur, and relative velocity which contribute to the visual quality of the formed CP system. Subsequently, we optimize throughput over energy consumption by adjusting transmission power and the number of utilized Radio Block (RB) by each vehicle, so that networking resource utilization is optimized under noisy channels with diverse conditions. It is worth mentioning that the selection process is dynamic due to the time-varying nature of the parameters. In other words, the optimal selection at time epoch $t_1$, may not necessarily remain optimal throughout the entire CP session $T$, due to the inherent dynamicity of CP quality factors. To account for this matter, we consider the time average over the CP interval, when solving the optimization problem. The resulting optimization problem, unfortunately, does not admit a convex optimization format for including fractional terms. To solve this problem, we first use Dinkelbach's algorithm to transform the original optimization problem into a series of parametric optimization problems, serving as surrogates for fractional terms.

Next, we apply Lagrangian relaxation and duality theory to decompose the problem into smaller sub-problems by relaxing the constraints and incorporating them into the objective function using Lagrange multipliers. The dual problem is then solved iteratively to optimize the primal variables while satisfying the relaxed constraints, ensuring an efficient and precise solution.
Once the CP is formed by selecting an optimal set of helper vehicles, the next step is to optimize networking resources across the selected vehicles under given networking conditions. To adopt more realistic assumptions, we follow the commonly adopted approach of modeling transmission errors in C-V2X for CP in terms of packet drop rate when the received signal strength drops below the sensing threshold as well as packet drops due to collisions \cite{gonzalez2018analytical,jeon2020explicit}. The CP uses only the surviving transmission packets. 
Since the resource optimization problem for selected vehicles does not admit convexity conditions, we proposed an algorithm that uses Dinkelbach's method to solve this problem, as we did for the vehicle selection problem. Next, we apply the Frank-Wolfe, a gradient-based optimization technique suited for constrained convex problems, to solve this problem. These methods, when coupled together, ensure a practical and systematic approach to addressing the optimization problem while satisfying the required constraints, as shown in our simulation results in section \ref{sec:simulations}. 
To investigate the effectiveness of the proposed approach, we conducted simulations to demonstrate how it employs the best helpers for the ego vehicle, taking into account factors such as visual range, location, and motion blur. To further enhance the CP performance, we then optimize the CP's overall throughput per energy consumption under channels with different noise levels. 
The results obtained from our two-step optimization process present notable improvements in CP by selecting the best helper to increase the ego vehicle's visual range. 
While we examined the performance of the proposed CP system for OD, it is generic and can be used for other applications, such as speed estimation, lane detection, vehicle classification, traffic sign detection, traffic light interpretation, etc.
The rest of the paper is organized as follows. In section \ref{sec:related_work}, a more inclusive review of related literature is presented. Section \ref{sec:system_model}, the details of the system model are presented. In section \ref{sec:performance}, performance metrics are discussed. Finally, the proposed method is evaluated through intensive simulations in section \ref{sec:simulations}, followed by concluding remarks in section \ref{sec:conclusions}.

\begin{figure}[H]
\centering
\begin{subfigure}{0.8\columnwidth}
  \includegraphics[width=\textwidth]{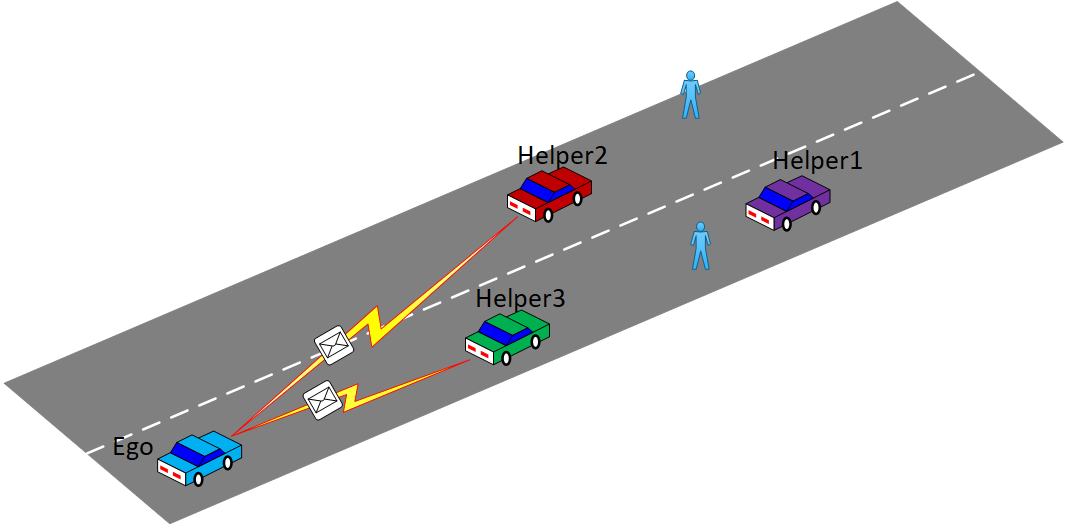}
  \caption{}
\end{subfigure}

\begin{subfigure}{1\columnwidth}
  \includegraphics[width=\textwidth]{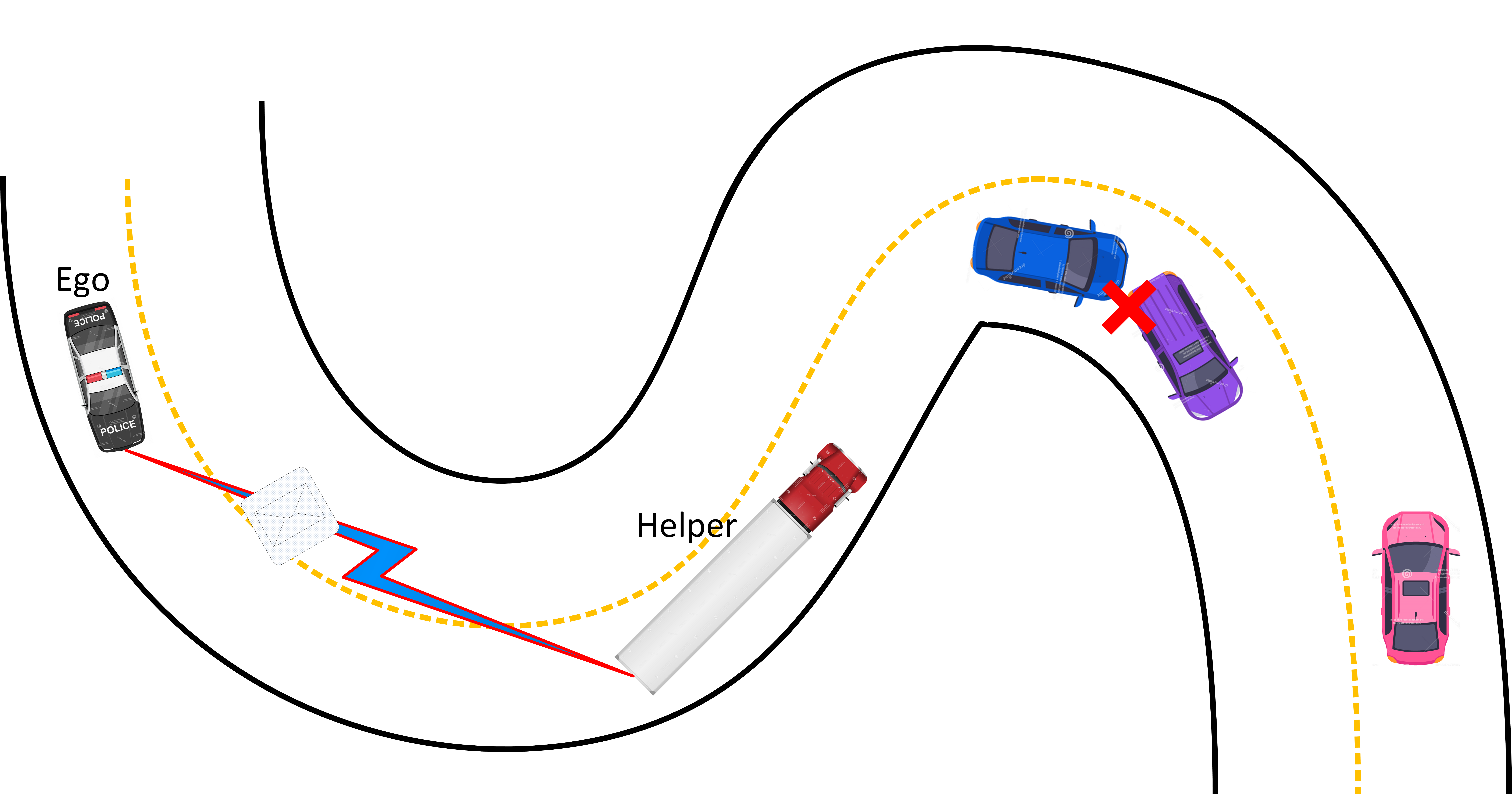}
  \caption{}
\end{subfigure}

\caption{Two scenarios where CP helps: (a) in foggy weather, the ego vehicle (blue) has poor perception quality, and two of the helper vehicles (green and red) share their camera feeds for improved pedestrian detection via CP, 
(b) on a winding road, the helper vehicle (truck) helps the ego vehicle (police car) detect the otherwise invisible accident.}
\label{fig:scenarios}
\vspace{-0.2 in}
\end{figure} 

\vspace{0.2 in}

\section{Related Work} \label{sec:related_work}

\subsection{CAVs under Imperfect Communications}
V2V LTE Release 14 (2017) presents advancements in V2V communication, enhancing connectivity and communication protocols for improved safety and efficiency in vehicular networks. However, there are technical limitations in V2V communications (like any other communication system), that should be considered when developing networked service provisioning systems. Similar requirements and limitations exist in newer versions of V2V, including 5G New Radio C-V2X Release 15 (2018) and Release 16 (2019). In this work, we chose the more robust version of Release 14 for its 
widespread use and availability of tools in network simulators.
A few prior studies have considered the impact of imperfect communication on the performance of service provisioning in the realm of vehicular communications. For instance, the authors of \cite{gonzalez2018analytical} evaluated the communication performance of C-V2X or LTE-V Mode 4, focusing on the average Packet Delivery Ratio (PDR) under four types of transmission errors, validating various transmission parameters and traffic densities.  Likewise,  \cite{chen2020joint, li2020federated} explored the impact of wireless factors, such as packet errors and limited bandwidth on the performance of Federated Learning (FL). Li et al. characterized the impact of efficient resource allocation techniques in V2V communication networks in FL, particularly for vehicular safety services, where delay and reliability requirements dictate an optimal utilization of limited spectrum resources \cite{li2022federated}. They emphasized that such resource allocation strategies are essential to mitigate interference, support various quality-of-service requirements, and ensure the success of emerging vehicle-related services in dynamic and fast-changing vehicular environments.
These research works, among others, suggest that imperfect communication can play a key role in CP; therefore, it should not be overlooked when forming a network of vehicles for collective perception. 
Hawlader et al. \cite{hawlader2024leveraging} investigates using edge and cloud computing to enhance real-time object detection in autonomous vehicles while balancing detection accuracy and latency. The authors explore offloading some computations to Multi-access Edge Computing (MEC) and cloud platforms. They create a synthetic dataset using the CARLA simulator to assess the trade-offs between prediction quality and delay. They conduct experiments using real hardware for inference and processing time measurements, supplemented with network simulations to estimate transmission latency. The study evaluates different variants of YOLOv5 models and explores the use of JPEG and H.265 compression techniques to optimize data transmission. Their findings demonstrate that integrating edge and cloud computing with optimized compression can enable real-time object detection at 20 Hz while maintaining high accuracy.

Authors in \cite{rashid2022energy} have explored energy-efficient resource allocation in multi-carrier NOMA-based heterogeneous networks (HetNets), addressing complex non-convex optimization challenges in user clustering, subchannel allocation, and power allocation. They utilized approaches such as suboptimal clustering algorithms and convexification techniques, including Dinkelbach's method, to enhance system energy efficiency while meeting QoS requirements and ensuring user fairness. The authors of \cite{jiang2021deployment} investigate the role of roadside units (RSUs) in providing traffic predictions on highways through vehicle-to-infrastructure (V2I) communication. The authors develop metrics to quantify the extent of traffic predictions available from RSUs and analyze how factors such as RSU placement, communication range, and connected vehicle penetration rate influence these predictions. Their findings demonstrate that even sparsely placed RSUs, up to 2000 meters apart, and low penetration rates of connected vehicles (as low as 2\%) can still yield meaningful traffic forecasts. The study proposes optimal RSU deployment strategies to maximize the efficiency of traffic prediction, offering valuable insights for designing connected roadside infrastructure.

\subsection{Object Detection} 
Our target application, in this work, is OD, which
involves identifying and locating objects in the pixel domain. 
It is an integral part of autonomous driving and crash avoidance \cite{li2023learning}. For instance, the trained 
RL algorithms naturally present sharp reactions to detected pedestrians to avoid fatal accidents.   
OD can be applied to different perception domains, including visions-based (e.g., regular and IR cameras), sensor-based (e.g., RADAR/LiDAR), and hybrid methods. OD entails detecting objects in three-dimensional space, especially when using LiDAR cloud points and imaging techniques. The following are noteworthy works in using OD out of many.   Chen et al. \cite{chen2024roadside} proposes an optimization approach for strategically placing Roadside LiDAR (RSL) sensors to enhance cooperative vehicle detection and tracking in Intelligent Transportation Systems. The authors develop a chance-constrained stochastic simulation-based optimization (SO) model to maximize the expected mean Average Precision (mAP) while ensuring a minimum recall level under traffic uncertainties. To address the computational complexity and black-box nature of the problem, they introduce a Gaussian Process Regression-based Approximate Knowledge Gradient (GPR-AKG) sampling algorithm. 
Reading et al. \cite{reading2021categorical} proposed Categorical Depth Distribution Network (CaDDN), as a monocular vision-based OD method. This method utilizes predicted categorical depth distributions for each pixel to enhance depth estimation accuracy and achieve a top-ranking performance on the KITTI dataset, showcasing its effectiveness in addressing the challenges of monocular detection. Authors of \cite{wang2022detr3d} introduced a multi-camera OD framework that operates directly in space, leveraging sparse object queries to index 2D features from multiple camera images.
PointRCNN is proposed in \cite{shi2019pointrcnn} 
as an OD framework that uses point cloud data points directly to generate high-quality proposals for OD, instead of generating proposals from RGB images or projecting point clouds to the bird's eye view.  Han et al. propose a hierarchical multi-lane platooning algorithm for Cooperative Automated Driving Systems (C-ADS) that integrates both strategic (mission-level) and tactical (motion-level) decision-making. The proposed approach enables organized multi-lane platooning behavior, including same-lane platooning, multi-lane joining, and on-ramp merging. At the tactical level, it generates predictive trajectories by incorporating intent-sharing among vehicles, ensuring safe and efficient maneuvers. The authors of \cite{yang2023cooperative} propose an Edge-empowered Cooperative Multi-camera Sensing (ECoMS) system for vehicle tracking and traffic surveillance using edge artificial intelligence and representation learning. It introduces a lightweight edge-based computer vision framework for real-time vehicle detection, tracking, and feature selection. It reduces bandwidth and computational costs by transmitting only object representations instead of full video streams. By leveraging edge computing and orchestrating data transmission, ECoMS efficiently models network-scale traffic information, offering a solution for intelligent transportation systems. Qi et al. \cite{qi2024geometric} propose a deep learning-based 3D object detection method, GIC-Net, designed to improve LiDAR-based perception for autonomous vehicles under adverse weather conditions like rain and snow. The method introduces a Geometric Location-Constrained (GLC) backbone module that incorporates geometric location information through ellipsoidal constraints to mitigate the impact of particle interference. Additionally, it employs a Line Geometric Feature Constraint (LGFC) module, which utilizes line features (such as length, slope, and intercept) to enhance object detection in sparse and incomplete point clouds.

An exemplary implementation of LiDAR-based OD methods is PointPillars, a highly effective point cloud encoder that employs PointNets to organize point clouds into vertical columns (pillars), demonstrating superior speed and accuracy \cite{lang2019pointpillars}. 
They outperform existing methods on the KITTI benchmarks while achieving higher accuracy in OD in comparison to \cite{yan2018second, xiang2017subcategory, ku2018joint}. Finally, a hybrid camera-LiDAR fusion detection method, called RCBEV, is proposed in \cite{zhou2023bridging}. RCBEV implements a radar-camera feature fusion method for OD in autonomous driving using nuScenes and VOD datasets through an efficient top-down feature representation and a two-stage fusion model that bridges the view disparity between radar and camera features.

Combining visual and LiDAR for enhanced visibility has been presented in \cite{xiang2024vidf, sindagi2019mvx, danapal2020sensor, ando1979generalized, mahmoud2023dense, huang2024semantics, li2022deepfusion, yoo20203d}. The majority of these multi-modal perception methods strive to combine visual feeds with RADAR and LiDAR readings to improve the perception quality of multi-input perception systems. While their primary focus is on single-vehicle scenarios, we emphasize multi-vehicle scenarios where the perception is rendered by a group of vehicles, bringing new challenges and requirements, such as optimal vehicle selection, and cooperation under imperfect communications, to be addressed in this work.
\vspace{-0.1cm}

\subsection{Fusion Methods}

The information of multiple perception systems can be collectively processed by merging information at three different levels, enabling \textit{early fusion}, \textit{integrative analysis}, and \textit{late fusion} \cite{sarlak2023cooperative, chu2025occlusion}.

In the \textit{early fusion} approach, raw information (r.g., camera feeds, sensor readings, LiDAR cloud points) are merged before further analysis. For instance, the authors of \cite{chen2019cooper} introduce an \textit{early fusion} approach enabling CP by combining LiDAR point clouds from diverse positions before processing them, which improves detection accuracy by expanding the effective sensing area. They validated their method using KITTI and T\&J datasets, demonstrating the feasibility of transmitting point cloud data via existing vehicular network technologies. 
Qiu et al. present an infrastructure-less CP system, leveraging direct vehicle-to-vehicle communication to efficiently share sensor readings while optimizing transmission schedules to enhance safety in dense traffic scenarios \cite{qiu2021autocast}. Implementing this method for visual information is more challenging because merging multi-view camera feeds in the pixel domain requires projecting all images into the same perspective. Wang et al. \cite{wang2024crrfnet} proposes an \textit{early fusion} method  CRRFNet. This method fuses camera and radar radio frequency (RF) data to enhance robustness and reliability in complex environments. Unlike traditional fusion techniques that rely on radar points, CRRFNet utilizes radar RF images, allowing for improved object classification even in challenging conditions such as sensor failures or severe interference. The model consists of specialized deep convolution modules for extracting features from both sensors, a deconvolution module for upsampling, and a heatmap module for compressing redundant channels, followed by location-based Non-Maximum Suppression (L-NMS) for object prediction. The paper also presents a new dataset, CRNJUST-v1.0, to complement the existing CRUW dataset for evaluating the method’s performance.

A \textit{late fusion} approach entails processing each perception signal individually and then combining the extracted semantic information. For instance, one may perform OD on individual camera feeds to obtain corner points, class labels, and confidence levels of detected objects, then fuse this information to obtain more refined high-fidelity results. An implementation of this method is proposed in \cite{lateef2023motion} for object identification (FOI) in urban driving scenarios, which extracts comprehensive object information (class, status, position, motion, and distance) from stereo camera images, eliminating the need for expensive sensor modalities like LiDAR. The framework integrates image registration for ego-motion compensation and optical flow estimation to detect moving objects and extract their behavioral features. A moving object detection model combines an encoder-decoder network with semantic segmentation to identify moving and static objects. The proposed system is evaluated on various datasets, including KITTI and EU-long-term, and demonstrates accurate semantic information extraction from a moving camera in complex environments.

The authors of \cite{rawashdeh2018collaborative} enhance the accuracy of shared information for collaborative automated driving. First, they employed Convolutional Neural Networks (CNN) for OD and classification to extract positional and dimensional information. Next, they combined extracted information to 
enhance driving safety 
in V2V collaborative systems. The authors in \cite{yu2022dair} introduce the DAIR-V2X dataset for Vehicle-Infrastructure Cooperative Autonomous Driving (VICAD) and formulate the Vehicle-Infrastructure Cooperative OD (VIC3D) problem, proposing the Time Compensation \textit{late fusion} framework as a benchmark, highlighting the importance of collaborative solutions and addressing challenges in autonomous driving. Article \cite{arnold2020cooperative} explores CP for OD using two fusion schemes, \textit{early} and \textit{late fusion}, showing that 
\textit{early fusion} outperforms \textit{late fusion} in challenging scenarios. Authors in \cite{yue2024wgs} introduce WGS-YOLO, an improved version of the YOLO object detection framework for autonomous driving. The model enhances feature fusion using the Weighted Efficient Layer Aggregation Network (W-ELAN) module with channel shuffling, which improves information integration. To optimize computational efficiency, the authors incorporate the Space-to-Depth Convolution (SPD-Conv) and Grouped Spatial Pyramid Pooling (GSPP) modules, reducing model complexity while maintaining high accuracy.

\textit{Early} and \textit{late fusion} methods offer alternative approaches by combining information either before or after applying the target DL-based application, like OD. This brings the convenience of deploying existing methods in the developed framework. Recently, some attempts have been made to develop integrative processing of multi-channel inputs to achieve elevated performance. 
Xu et al. \cite{xu2022v2x} introduced a fusion model V2X-ViT, based on the popular vision transformer architecture equipped with heterogeneous multi-agent and multi-scale window self-attention modules for robust CP in AVs that utilize V2X communication. They reported state-of-the-art (SOTA) OD performance in challenging and noisy environments using a large-scale V2X perception dataset. 
Another \textit{integrative} end-to-end DL-based architecture is presented in \cite{cui2022coopernaut} to implement an end-to-end learning model for vision-based cooperative driving. They utilized cross-vehicle perception with LiDAR data encoded into compact point-based representations, demonstrating an improvement in average success rate over egocentric driving models in challenging scenarios, with a 5× smaller bandwidth requirement compared to prior work, as evaluated in the AUTOCASTSIM simulation framework. 
An important challenge of these approaches is their custom-built natures, complicating integration with current and upcoming versions of DL architectures that are primarily designed for single perception systems. In this work, we adopt a \textit{late fusion} approach for its convenience and easy integration with SOTA DL methods, to demonstrate the performance of CP with optimal vehicle selection and networking resource allocation under imperfect communication.

\section{System Model} \label{sec:system_model}

In the proposed framework, the ego vehicle sends request-for-help messages to front vehicles with potentially better visions, to initiate the CP session when individual perception quality is not satisfactory. A summary of notations is listed in Table \ref{tab:notations}. 
We assume that  
$N$ vehicles, $V_1$, $V_2$, \dots, $V_N$, notify their intention to join the CP session through Ack messages along with their position ($x_i$, $y_i$), speed $v_i$, channel conditions, and sample images. Additional details, such as the distance to the ego vehicle $x
$, approximate vision range $R$, motion blur, field of view, required transmission energy $\mathbb{E}$, and communication errors in the C-V2X $\delta_{\text{Er}}$, are estimated based on the gathered information (position, channel conditions) and the collected sample images. For instance, motion blur is directly related to the vehicle velocity \cite{xiao2021vehicle}.  

\begin{table}[h!]
\centering
\caption{List of Notations and Definitions}
\begin{tabular}{|c|l|}
\hline
\textbf{Notation} & \textbf{Definition} \\ \hline
$\rho$ & Density in PPP \\ \hline
$\mu$ & Mean in PPP \\ \hline
$V_i$ & Vehicle $i$ \\ \hline
$x$ & Position in the $x$-axis \\ \hline
$y$ & Position in the $y$-axis \\ \hline
$v$ & Velocity \\ \hline
$t$ & Time \\ \hline
$\text{R}$ & Visual Range \\ \hline
$e, z, u, r$ & Camera Parameters \\ \hline

$s_i$ & Volunteer Vehicle $i$ \\ \hline
$T$ & Time Interval \\ \hline
$N$ & Number of Helpers \\ \hline
$M$ & Number of Helper Selection \\ \hline
$W_{\text{subCh}}$ & Number of Subchannel \\ \hline
$P_{\text{rb}}$ & Probability of Choosing a Particular RB  \\ \hline
$w_T$ & Total Number of Available RBs \\ \hline
$w_i$ & Number of RBs allocated to vehicle $i$\\ \hline
$P_i^{\text{rx}}$ & Received Signal Power \\ \hline
$P_i^{\text{tx}}$ & Transmit Power \\ \hline
$d_{ij}$ & Distance between vehicle $i$ and $j$ \\ \hline
$SH$ & Shadowing \\ \hline
$\gamma$ & Path Loss Exponent \\ \hline
$\delta_{\text{COL}}$ & Collision Error \\ \hline
$\delta_{\text{SEN}}$ & Transmission Error \\ \hline
$\mathbb{E}$ & Energy Consumption\\ \hline
$\zeta$ & Throughput of the Channel\\ \hline
\end{tabular}
\label{tab:notations}
\end{table}

\begin{figure}[htbp]
\begin{center}
\centerline{
\includegraphics[width=0.5\columnwidth]{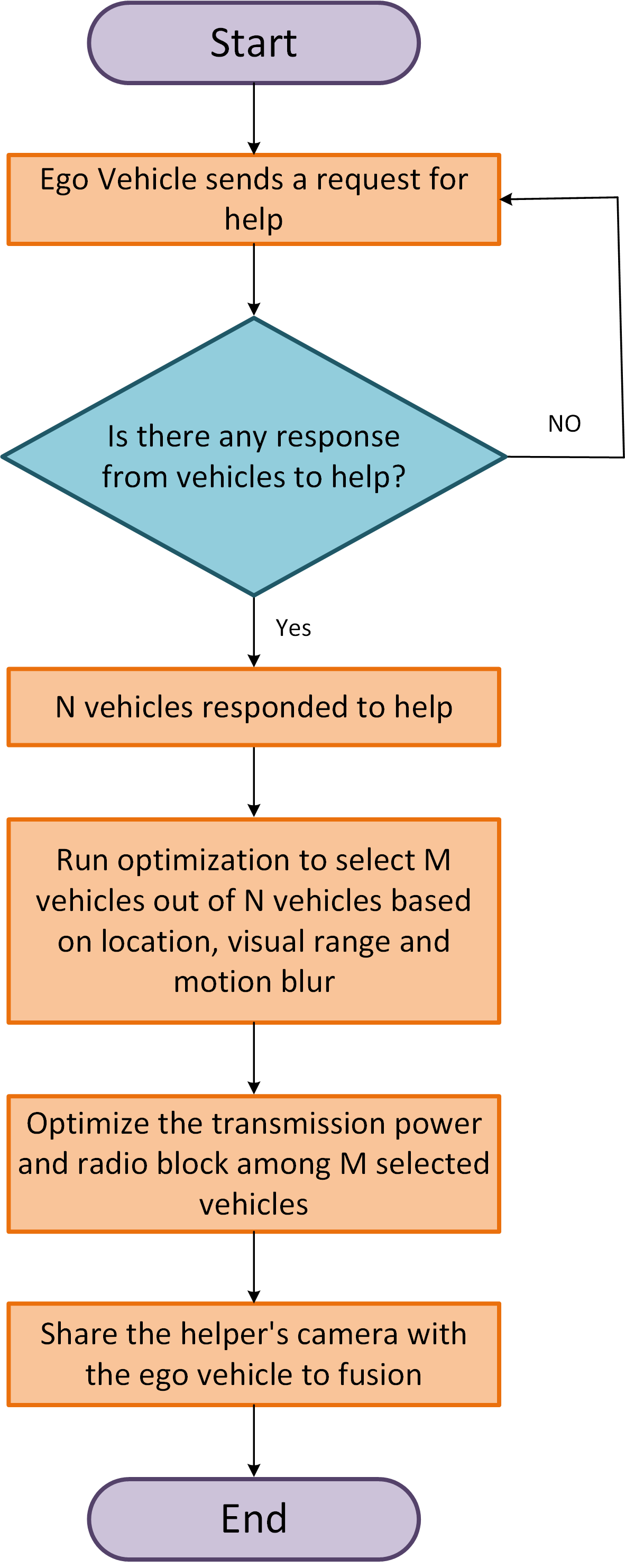}}
\vspace{-0.1cm}
\caption{The workflow of multi-vehicle CP implementation.}
\label{fig:fl1}
\end{center}
\end{figure}

Once the information is obtained, the ego vehicle runs an optimization problem (\ref{eq:objective_function}) and selects $M$ out of $N$ volunteered vehicles that can collectively achieve the highest performance under given conditions. Fig.\ref{fig:fl1} represents the sketch of this process.

The main objective of selecting helper vehicles is to maximize the overall perception quality 
through a practical and informed choice of helpers. In our case, the key performance metrics include the vehicle's contribution to extending the visual range of the ego vehicle (based on their positions) and their image quality based on motion blur (based on their velocities), while maintaining the communication cost as low as possible. 

In order to model the helper vehicle locations, we use the commonly adopted Poisson Point Process (PPP) for random arrivals. Specifically, we consider a PPP with density $\rho$ for the number of vehicles (helpers) on the highway over an interval \([a, b]\) with mean $\mu = \rho (b - a)$.

This is equivalent to assuming an exponential distribution $ \rho e^{-\rho x} \quad \text{for} \quad x \geq 0$ for inetr-arrival time $x$.

We use a Truncated Gaussian distribution for vehicle velocities which best mimics the velocity variation based on experimental data, while embracing the fact that velocity \( v \) restricted to the interval \([v_{\text{min}}, v_{\text{max}}]\) \cite{yousefi2008analytical}. Specifically, when limiting normally distributed $v\sim N(\mu, \sigma^2)$ to interval $[v_{\text{min}}, v_{\text{max}}]$, we obtain $f(v \mid v_{\text{min}} \leq v \leq v_{\text{max}}) = \frac{1}{\sigma \sqrt{2\pi} \big(\phi(\beta) - \phi(\alpha)\big)} e^{-\frac{(v - \mu)^2}{2\sigma^2}}$, where, $\alpha = \frac{v_{\text{min}} - \mu}{\sigma},
\beta = \frac{v_{\text{max}} - \mu}{\sigma},
Z = \frac{v - \mu}{\sigma}$ and $\phi(.)$ is the Cumulative Distribution Function (CDF) of the standard normal distribution. To analyze the positional dynamics of vehicles over time, it is essential to first assess whether or not the contributing terms form a stationary process. A non-stationary process indicates variables vary over time, undermining the ego vehicle's attempt to identify and select suitable helper vehicles. A stationary process is a stochastic process whose unconditional joint probability distribution does not change when shifted in time. Consequently, parameters such as mean and variance also do not change over time.

A continuous-time random process \(\{Z(t), t \in \mathbb{R}\}\) is strict-sense stationary or simply stationary if, for all \(t_1, t_2, \dots, t_n \in \mathbb{R}\) and all \(\Delta \in \mathbb{R}\), the joint CDF of \(Z(t_1), Z(t_2), \dots, Z(t_n)\) is the same as the joint CDF of \(Z(t_1 + \Delta), Z(t_2 + \Delta), \dots, Z(t_n + \Delta)\). That is, for all real numbers \(z_1, z_2, \dots, z_n\), we have

\begin{align}
&F_{Z(t_1)X(t_2)\dots Z(t_n)}(z_1, z_2, \dots, z_n)  \\ \nonumber
&~~~~= F_{Z(t_1 + \Delta)Z(t_2 + \Delta)\dots Z(t_n + \Delta)}(z_1, z_2, \dots, z_n).
\end{align}

Since the optimal vehicle selection relies on the visual range of candidate vehicles, which itself is a function of their relative distances, it is crucial to ensure that the relative distances remain stable over time. This stability allows the results of optimization, computed at a single time point, to remain valid throughout the CP interval.
To formalize this, we define the stochastic process, as
\begin{align}
    \Psi = \{Y_1(t), Y_2(t), \ldots, Y_{N-1}(t)\}, 
\end{align}
\noindent where each $Y_i(t)$ represents the relative distance between vehicle $i$ and its preceding vehicle $i+1$, defined as
\begin{align}
    Y_i(t) &= X_{i+1}(t) - X_i(t)\\ \nonumber
    &= t(v_{i+1} - v_i) + \text{constant}
\end{align}
where $X_i(t)$ and $V_i(t)$ are the position and velocity of vehicle $i$ at time $t$.

For the optimization results to hold across the entire CP session, the process $\Psi(t)$ must remain stationary. In other words, the statistical properties of $\Psi(t)$, such as the mean and variance of $Y_i(t)$, should remain constant over time. If $\Psi(t)$ is stationary, the optimization based on a snapshot of vehicle positions at a single time point is valid for the entire CP interval. A necessary condition for $\Psi(t)$ being stationary is $E[v_{i+1}(t) - v_i(t)]=0$, which implies equal expected velocity for all vehicles. Under constant velocity assumption (no acceleration/deceleration), it simplifies to equal velocity for all vehicles (i.e. $v_1=v_2=...v_N$). This assumption is restrictive and trivial.  
Therefore, $\Psi(t)$ is non-stationary in most realistic scenarios, meaning the relative distances vary significantly over time. In such cases, time-averaging should be considered for the objective functions during the entire CP, session to account for the dynamic behavior of the system (e.g., equation\ref{eq:optimization}).

Another consideration is addressing imperfect communication between vehicles, since the errors in the wireless network can severely reduce the perception quality by processing noisy, partially masked, or intermittent images. To mitigate this issue, it is essential to efficiently allocate transmission resources (e.g., RBs and transmission power) to minimize transmission errors while ensuring efficient utilization of the communication channel.

In LTE 4G communication, RBs of time and frequency are structured to ensure efficient data transmission. Time is divided into frames of 10 milliseconds (ms), each containing ten subframes lasting 1 ms, split into two timeslots of 0.5 ms. On the frequency side, the bandwidth is divided across RBs, each consisting of 12 subcarriers spaced by 15 kHz and holding 14 OFDM symbols. If a vehicle is allocated a 10 MHz channel, it could be assigned up to 50 RBs, where each RB occupies 180 kHz.
Scheduling plays a crucial role through the RB Reservation Interval (RRI), which dictates when vehicles utilize reserved RBs. The subchannel size can vary between 4 to 50 RBPs, and a vehicle may be assigned multiple subchannels within a subframe to transmit data effectively.

In NR-V2X mode 2, vehicles autonomously select RBs from sideline configurations provided by the gNB or pre-configured settings. To enhance the semi-Persistent Scheduling (SPS) algorithm from LTE-V2X mode 4, 3GPP proposes modifications to support periodic and non-periodic traffic with flexible packet sizes for advanced safety services \cite{lusvarghi2023comparative, sehla2022resource}. 

There exist two channel sensing mechanisms, including long-term sensing and short-term sensing. The former is inherited from LTE-V2X for predictable, periodic traffic. The short-term sensing is based on listen-before-talk (LBT), which is more suitable for nonperiodic, unpredictable traffic. In CP, the traffic is predominantly periodic, since vehicles need to share sensor data and awareness messages regularly to maintain a consistent and up-to-date understanding of the environment. This periodic exchange is critical for maintaining synchronized knowledge of objects such as pedestrians, obstacles, or nearby vehicles to support applications like autonomous driving and platoon management. However, nonperiodic traffic can also occur in response to unexpected events, such as sudden braking or accident alerts, which require immediate, ad-hoc communication, which is not part of our work. 
NR-V2X integrates both sensing mechanisms to balance efficiency and reliability, ensuring that regular status updates and urgent event-driven messages coexist seamlessly, supporting a robust and low-latency communication framework for future intelligent transportation systems. In our work, we use long-term sensing for periodic CP updates. 

The SPS algorithm enables vehicles to autonomously select RBs from a pre-configured pool by sensing the channel to identify unused RBs. Vehicles reserve RB for a series of transmissions, guided by the reselection counter, which controls how many messages are sent before a new RB selection is required. Essential information, such as the periodicity of Cooperative Awareness Messages (CAM) and the counter value, is shared via Sidelink Control Information (SCI) to prevent RB collisions. The selection process involves filtering candidate RBs using metrics like signal power and channel occupancy, with two lists, $L_1$ and $L_2$, created to refine available RBs. In the end, when the SPS algorithm finishes, there are available RBs in the list $L_2$, and vehicles should select the RBs randomly from the list, which still there is a high chance for collision, especially when cars at the same time need numbers of RB to transmit the data. We consider collision-related errors in our implementation to make it more realistic, which applies to both LTE-based and NR-5G V2X communications.

\begin{figure}[htbp]
\begin{center}
\centerline{
\includegraphics[width=1\columnwidth]{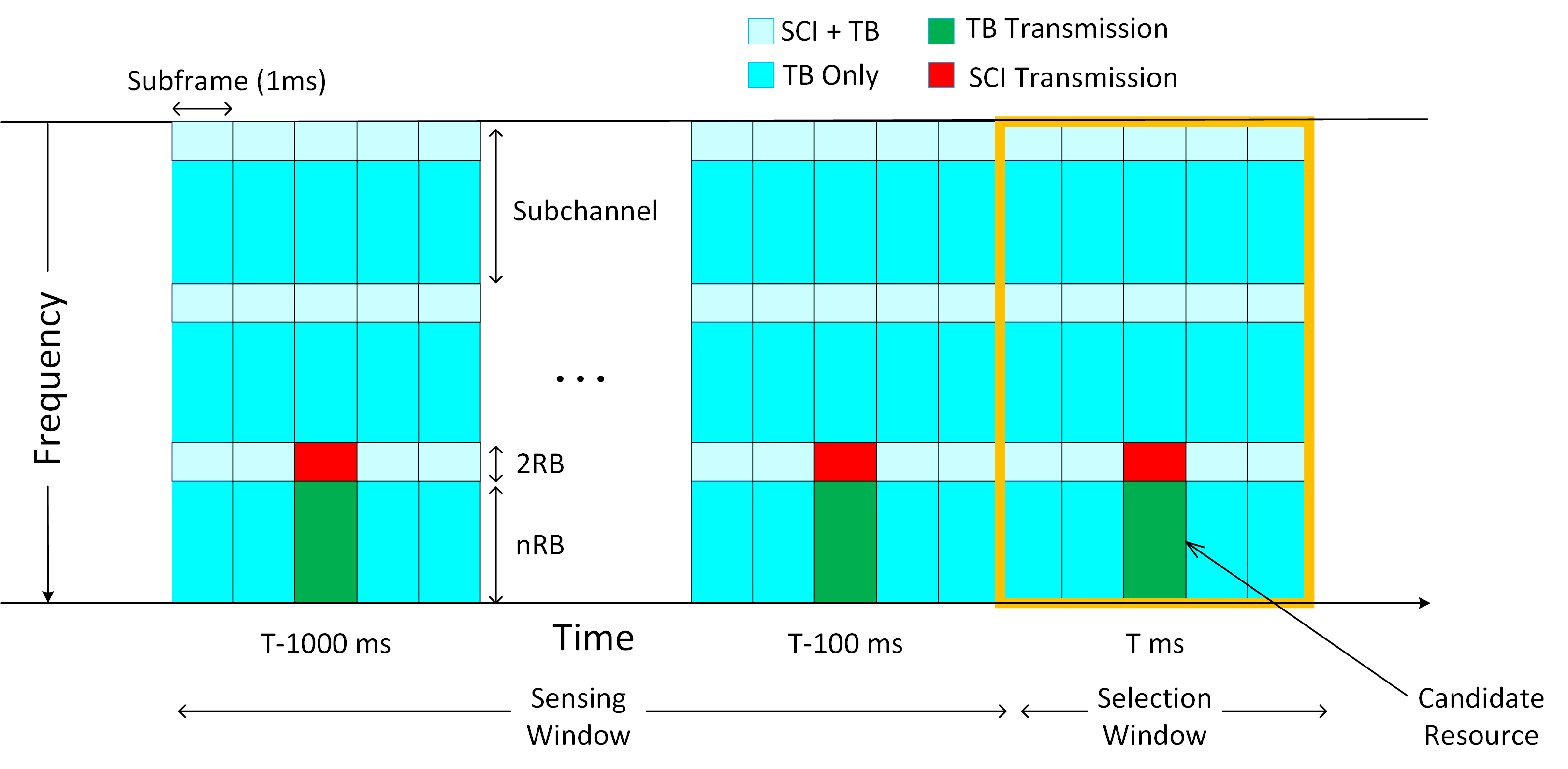}}
\vspace{-0.1cm}
\caption{C-V2X subchannelization and example of utilization resources. Resource contain of Transport Block (TB) and SCI.}
\label{fig:rb1}
\end{center}
\end{figure} 

In our problem, we assume that once the vehicles are selected, a set of RBs are available, as illustrated in Fig. 4. These RBs are utilized by the selected vehicles to transmit data efficiently, ensuring effective power transmission. This approach aims to minimize energy consumption and decrease the error rate within the network.

\section{Performance Metrics}\label{sec:performance}

\subsection{Location}
\begin{figure*}[htb]
    \centering
    \centerline{\includegraphics[width=1\textwidth]{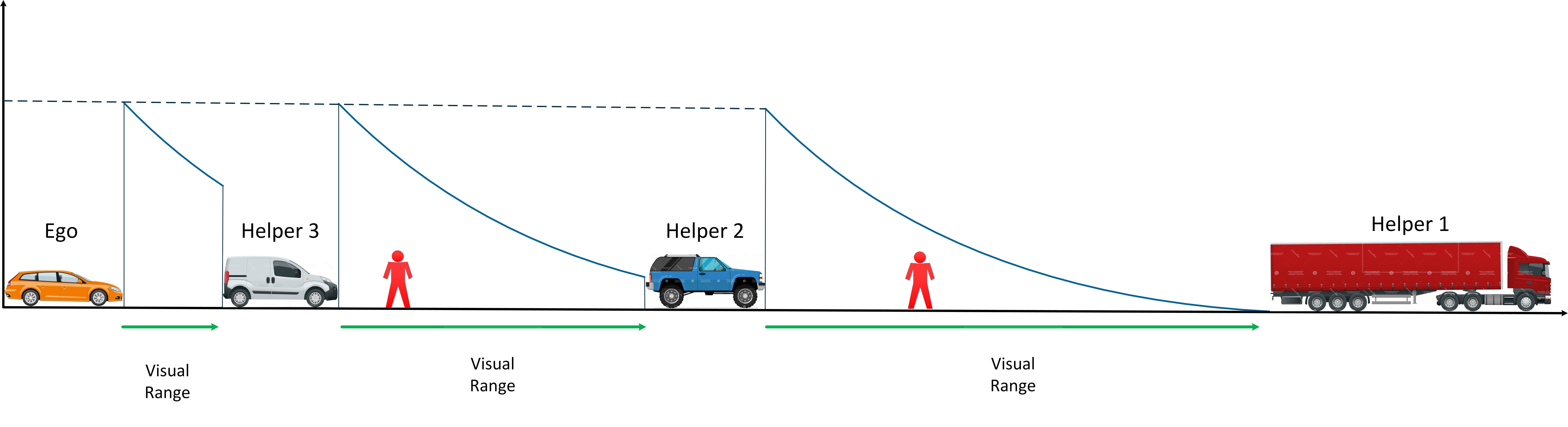}}
    \caption{The individual visual range for vehicles. Recruiting helpers extend the visual range of the ego vehicle through CP.}
    \label{fig:vs1}
\end{figure*}

Prioritizing the selection of nearby vehicles is critical to ensure accurate detection of proximate objects, which is vital for safe driving operations. The helpers' relative positions play a key role in achieving a reliable perception of the surrounding environment. For instance, as illustrated in Fig. \ref{fig:vs1}, Helper 3 contributes to the detection of both pedestrians, either fully or partially, whereas Helper 2 assists in detecting only one pedestrian, and Helper 1 provides no relevant contribution to the detection of either pedestrian. Also, detecting pedestrians closer to the ego vehicle has apparent advantages for enabling timely reactions. 

Let the binary vector $\mathbf{s} = (s_1, s_2, \ldots, s_N)$ denote the set of volunteer vehicles, where $s_i = 1$ indicates that vehicle $i$ is selected, and $s_i$ = 0 indicates that it is not selected. The weighted augmented position of the selected vehicles can be mathematically expressed as:

\begin{equation}
    f_1(\mathbf{s}) = \sum_{i=1}^{N} \sum_{t=0}^{T} s_{i} (x_{it} - x_{0t}), \label{eq:f1}
\end{equation}

where

\begin{equation} \nonumber x_{it} = x_{i0} + v_{i}t. \end{equation}

Here, $x_{it}$ represents the position of helper $V_i$ at time $t$, $v_i$ is the velocity of vehicle $V_i$ ($V_0$ is the ego vehicle). During the $T$ time interval, it is assumed that the velocity of each vehicle remains constant.

\subsection{Visual Range}
An essential factor for selecting helper vehicles is the enhancement of the ego vehicle's collective visual range, which directly impacts the accuracy of OD. The visual range of each vehicle is influenced by factors such as weather conditions, ambient lighting, and the presence of obstacles in the environment. Fig. \ref{fig:vs1} illustrates individual vehicles' effective visual range.
To maximize the collective visual range of the ego vehicle, the selection process must be optimized to ensure that the chosen helpers significantly contribute to expanding this range. All vehicles are assumed to know their respective positions, $(x_i, y_i)$ and their distances to the front vehicle, and can share this information with the ego vehicle.
The visual range function is mathematically defined as:
\begin{equation}
    f_2(\mathbf{s}) = \frac{1}{\sum_{i=1}^{N} \sum_{t=0}^{T} s_{i} \text{R}_{it}}, \label{eq:f2}
\end{equation}
where
\begin{equation}  \nonumber \text{R}_{it} = x_{(i+1)t} - x_{it}. \end{equation}
Here, $\text{R}_{it}$ denotes the visual range of vehicle $i$ at time $t$, defined as the distance between vehicle $i$ and its subsequent vehicle $i+1$ at time $t$. By leveraging this formulation, the optimization process ensures that the selected vehicles maximize the overall effective visual range of the ego vehicle. 

\subsection{Motion Blur}
We assume that all vehicles are equipped with 
similar cameras with equal resolution and field of view. In this paper, the only important parameter affecting motion blur is the speed of the vehicles, as  
\begin{equation} 
    f_3(\mathbf{s}) = \sum_{i=1}^{N} \sum_{t=0}^{T} s_i \left( \frac{v_{i} e \left[ r \cos(\varphi) - uQ \sin(\varphi) \right]}{v_{i} eu \sin(\varphi) + zu} \right).
\end{equation}

For the vehicles joining CP ($s_i=1$), $e$ is the exposure time interval, $r$ is the camera focal length, $u$ is the charge-coupled device (CCD) pixel size in the horizontal direction, $Q$ is the starting position of the object in the image (in pixels), $\varphi$ is the angle between the motion direction and the image plane, $z$ is the perpendicular distance from the starting point of the moving object to the pinhole \cite{cortes2018velocity}, also, $s_i$ is the selection variable for vehicle $i$, $v_{it}$ is the velocity of vehicle $i$ at time $t$.

The effect is maximized and becomes linear in velocity, when the image plane is parallel to the motion direction ($\varphi = 0$), as we get 

\begin{equation}
    f_3(\mathbf{s}) = \sum_{i=1}^{N} \sum_{t=0}^{T} s_i \left( \frac{v_{it} er}{zu} \right) \label{eq:f3}
\end{equation}

\subsection{Optimization Problem}
In scenarios involving vehicle location, visual range, and motion blur, an optimization problem emerges to select helper vehicles. The objective is to identify vehicles that maximize the overall visual coverage for the ego vehicle, while simultaneously enhancing image quality by minimizing motion blur and prioritizing closer vehicles.

The goal is to select the best $M$ vehicles out of a total of $N$ candidates and allocate resources among the selected vehicles. The optimization process is governed by transmission Key Performance Indicators (KPIs) calculated for the selected vehicles. Let the binary vector $\mathbf{s}=[s_1,s_2,\dots,s_N]^T$ represent the selection status of vehicles, where $s_i=1$ if vehicle $i$ is selected and $s_i=0$ otherwise.

The optimization problem is formulated as:

\begin{align}
     \text{minimize} \quad &\sum_{i=1}^{N} \sum_{t=0}^{T} f(\mathbf{s}), \nonumber\\ 
     \text{subject to:}  \quad &s_i \in {0,1}, \quad i = 1, \dots, N, \nonumber \\  
     &\sum_{i=1}^{N} s_i \leq M, \quad i = 1, \dots, N, \label{eq:optimization}
\end{align}
\noindent where $f(\mathbf{s})$ is a composite objective function defined as:

\begin{equation} f(\mathbf{s}) = f_1(\mathbf{s}) + f_2(\mathbf{s}) + f_3(\mathbf{s}), \end{equation}
\noindent with $f_1(\mathbf{s})$, $f_2(\mathbf{s})$, and $f_3(\mathbf{s})$ capturing distinct components of the optimization criteria, such as location, visual coverage, and motion blur reduction enhancement.

The function $f(\mathbf{s})$ is expressed in its fractional form for computational efficiency and clarity:

\begin{equation}  f(\mathbf{s}) = \frac{G(\mathbf{s})}{D(\mathbf{s})}, \end{equation}

\noindent where

\begin{align}  G(\mathbf{s}) &= (\mathbf{s}^T \cdot \mathbf{x} * \mathbf{R}^T \cdot \mathbf{s})zu \nonumber \\
&+ zu + (\mathbf{s}^T \cdot \mathbf{v} \star \mathbf{R}^T \cdot \mathbf{s})er,  \label{eq:Formul1}
\end{align}  

\begin{align}
    D(\mathbf{s}) &= zu . (\mathbf{R}^T * \mathbf{s}). \label{eq:Formul2}
\end{align}  

\noindent where $ \mathbf{R} $ and $ \mathbf{x} $ are \( N \times T \) matrices,
 \( \mathbf{v} \) is a vector,
 \( z, u, e, \) and \( r \) are scalar values,
As we know this is a binary and fractional problem which
is not convex.
Here, $\mathbf{x}$ and $\mathbf{v}$ represent position matrix and velocity vector, respectively, $\mathbf{R}$ denotes the visual range matrix. By solving this optimization problem, the selected vehicles effectively maximize the ego vehicle's visual coverage while meeting the transmission KPIs.

The fractional programming problem can be formulated as follows:
\begin{equation}
    \min \left\{ \frac{G(\mathbf{s})}{D(\mathbf{s})} \right\}
\end{equation}

where \( G(\mathbf{s}) \) and \( D(\mathbf{s}) \) represent the numerator and denominator of the fractional objective function, respectively.

We can apply the Dinkelbach algorithm \cite{rodenas1999extensions} to transform the fractional nature of the objective function into a standard optimization form.

\textbf{Lemma \cite{rodenas1999extensions}:} The function \( f(\eta) = \min \left\{ G(\mathbf{s}) - \eta D(\mathbf{s}) \right\} = 0 \) if and only if 
\begin{align} 
\eta = \frac{G(\mathbf{s}^*)}{D(\mathbf{s}^*)} = \min \left\{ \frac{G(\mathbf{s})}{D(\mathbf{s})} \right\}
\end{align}

\textbf{Proof:} See \cite{rodenas1999extensions}.

The Dinkelbach algorithm can be outlined as follows.

\textbf{Step 1:} Choose an arbitrary \( \chi_{1} \in \mathbb{S} \) and set \( \eta_2 = {G(\chi_{1})}/{D(\chi_{1})} \) (or set \( \eta_2 = 0 \)).

\textbf{Step 2:} Solve the problem 
\begin{align} 
F(\eta_k) = \min \{ G(\chi) - \eta_k D(\chi) \mid \chi \in \mathbb{S} \}.
\end{align}

After solving the problem, denote the optimal solution as \( \chi_k \).

\textbf{Step 3:} 
If \( F(\eta_k) < \epsilon \) (optimality tolerance), stop and output \( \chi_k \) as the optimal solution. If \( F(\eta_k) \geq \epsilon \), let \( \eta_{k+1} ={G(\chi_k)}/{D(\chi_k)} \) and go to Step 2, replacing \( k \) with \( k + 1 \) and \( \eta_k \) with \( \eta_{k+1} \).

\textbf{Proposition:} \( F(\eta^*) = \min \{ G(\chi) - \eta^* D(\chi) \mid \chi \in \mathbb{S} \} = 0 \) if and only if \( \eta^* = {G(\chi^*)}/{D(\chi^*)} = \min \{{G(\chi)}/{D(\chi)} \mid \chi \in \mathbb{S} \} \).

Plugging-in our values from (\ref{eq:Formul1}) and  (\ref{eq:Formul2}) in the objective function of our optimization problem in (\ref{eq:optimization}), then using DinkelBach's method, yields the following optimization problem.

\begin{align}
     \text{minimize} & \big[ \mathbf{s}^T \left(\mathbf{x} * \mathbf{R}^T zu + \mathbf{v} \star \mathbf{R}^T er \right) \mathbf{s} \nonumber \\ 
     &~~~~~~~- \eta (zu\mathbf{R}^T)\mathbf{s}+ zu \big]   \nonumber \\ 
     \text{subject to:}  \quad &s_i \in {0,1}, \quad i = 1, \dots, N, \nonumber \\  
     &\sum_{i=1}^{N} s_i \leq M, \quad i = 1, \dots, N, \label{eq:optimization1}
\end{align}

\noindent which should be solved iteratively.
This problem can be cast as a Nonconvex Quadratically Constrained Quadratic Program (QCQP). To develop a more compact form, let's define the following matrices and vectors:

\begin{align} 
\nonumber
    P_0 &= \left(\mathbf{x}*\mathbf{R}^T zu + \mathbf{v} \star \mathbf{R}^T er\right)\\
\nonumber
q_0 &= \eta \left( zu \mathbf{R}^T \right)\\
    h_0 &= zu
\end{align}

With these definitions, the objective function converts to:

\begin{align}
    \operatorname*{minimize} \left\{ \mathbf{s}^T P_0 \mathbf{s} + q_0^T \mathbf{s} + h_0 \right\}.
\end{align}

QCQPs generalize the classic Quadratic Programming (QP) problems by permitting constraints to involve second-order polynomial terms. QCQPs are notably more complex than the standard QP problems. Specifically, nonconvex QCQPs are characterized by constraints or objective functions that are not convex, complicating the solution process due to the potential presence of multiple local minima.
Mathematically, a general nonconvex QCQP can be formulated as follows:

\begin{align} 
\nonumber
    \text{minimize} \quad &s^T P_0 \mathbf{s} + q_0^T \mathbf{s} + h_0 \\
    \text{subject to} \quad &s^T P_i \mathbf{s} + q_i^T \mathbf{s} + h_i \leq 0, \quad i = 1, \dots, m \label{eq:optimization2}
\end{align}

Here, the variable $\mathbf{s}$  belongs to \( \mathbb{R}^n \), and the matrices and vectors are defined as follows: \( P_i \in \mathbb{S}^n \) (the space of \( n \times n \) symmetric matrices), \( q_i \in \mathbb{R}^n \), \( h_i \in \mathbb{R} \).

If all matrices \( P_i \) are positive semidefinite, the problem is convex and can be solved efficiently using standard convex optimization techniques. However, if at least one of the matrices \( P_i \) is not positive semidefinite, the problem becomes nonconvex. In such cases, the QCQP is NP-hard, meaning that finding an optimal solution is computationally challenging and generally requires sophisticated methods or heuristics \cite{d2003relaxations, boyd2004convex}.


Our problem \ref{eq:optimization1} can be represented compactly as:

\begin{align}
    \text{minimize} \quad &\sum_{i=1}^{N} \sum_{t=0}^{T} \left( \mathbf{s}^T P_0 \mathbf{s} + q_0^T \mathbf{s} + h_0 \right)   \nonumber
     \\
    \text{subject to} \quad &s_i \in \{0, 1\}, \quad i = 1, \dots, N, \nonumber \\
    &\sum_{i=1}^{N} s_i \leq M, \quad i = 1, 
\end{align} 

In order to turn it into the standard form of QCQP in \ref{eq:optimization2}, we manipulate the constraint functions as follows.

The constraint $\sum_{i=1}^{N} s_i \leq M, \quad i = 1, \nonumber \dots, N$ can be put in the 
$s^T P_is + q_i^T s + h_i \leq 0$ by setting

\begin{align}
 P_i = 0 ,  q_i = I^T , h_i = -M. 
\end{align}

Likewise, constraints $s_i \in \{0, 1\}, \quad i = 1, \dots, N$ can be converted to QCQP constrain format $s^T P_is + q_i^T s + h_i \leq 0$, by setting

\begin{align}
 P_i = \mathbf{I}_{N\times N},  q_i = \mathbf{1}, 
\end{align}
where $\mathbf{I}$ and $\mathbf{1}$ are the identity matrix and all-one vectors. 
This is due to the fact that $ s_i \in \{0, 1\} $ are the only solutions for 
\begin{align} 
   \mathbf{s}^T \mathbf{I} \mathbf{s} - \mathbf{1}^T \mathbf{s}= 0
\end{align}
equivalent to 
\begin{align} 
   s_i^2 - s_i = 0, \text{ for } i = 1,2,\cdots,N.
\end{align}

Therefore, our optimization problem can be represented as 
\begin{align} \label{eq:objective_function}
    \nonumber
    \text{minimize} \quad  &\mathbf{s}^T P_0 \mathbf{s} + q_0^T \mathbf{s} + h_0 \\
    \nonumber
    \text{subject to}  \quad &q_i^T \mathbf{s} - M \leq 0,\\
    &\mathbf{s}^T \mathbf{I} \mathbf{s} - \mathbf{1}^T \mathbf{s} = 0.
\end{align}

As mentioned earlier, this problem does not admit a closed-form solution. To solve it, we employ the Lagrangian relaxation method to derive a computationally efficient lower bound on the optimal value of the nonconvex QCQP. This approach leverages the fundamental property that the dual formulation of any optimization problem is always convex, irrespective of the convexity of the primal problem, and is thus prone to efficient solution techniques. The Lagrangian relaxation method begins by transforming the original nonconvex QCQP into its dual representation, where the nonconvex constraints are incorporated into the objective function using Lagrange multipliers. This transformation enables the decoupling of the constraints, thereby simplifying the problem into a more tractable form. Specifically, starting from the nonconvex QCQP, the Lagrangian function is formulated as:

\begin{align} 
L(\mathbf{s}, \lambda) &= \, \mathbf{s}^T P_0 \mathbf{s} + \mathbf{s}^T \left( \sum_{i=1}^{N} \lambda_i P_i \right) \mathbf{s} \nonumber \\ \nonumber
&+ \left( q_0 + \sum_{i=1}^{N} \lambda_i q_i \right)^T \mathbf{s} \\ 
&+ h_0 - \sum_{i=1}^{N} \lambda_i M.
\end{align}

To find the dual function, we minimize it over $\mathbf{s}$ using the general formula

\begin{align}  
\inf_{\mathbf{s} \in \mathbb{R}^n} \left( \mathbf{s}^T P(\lambda) \mathbf{s} + q(\lambda)^T \mathbf{s} + h(\lambda) \right) 
\end{align}

\noindent with
\begin{align} 
    P(\lambda) = P_0 + \sum_{i=1}^{m} \lambda_i P_i, \\
    q(\lambda) = q_0 + \sum_{i=1}^{m} \lambda_i q_i,  \\
    r(\lambda) = h_0 + \sum_{i=1}^{m} \lambda_i h_i.
\end{align}

It is possible to derive an expression for \( g(\lambda) \) for general \( \lambda \), but it is quite complicated. If \( \lambda \geq 0 \), however, we have \( P(\lambda) \succ 0 \) and only may et the gradient to zero to minimize $L$ over $\mathbf{s}$, as follows 
\begin{align} 
    \nabla_s L(\mathbf{s}, \lambda) &= 2 P(\lambda)     \mathbf{s} + q(\lambda) = 0 \nonumber \\ 
    &\implies P(\lambda) \mathbf{s} = -\frac{1}{2} q(\lambda) \nonumber \\ 
&\implies \mathbf{s} = -\frac{1}{2} P(\lambda)^{\dagger} q(\lambda).
\end{align}

plug $\mathbf{s}$ into $L$ to obtain 

\begin{align} 
    g(\lambda) &= L\left(-\frac{1}{2} P(\lambda)^{\dagger} q(\lambda), \lambda\right) \\ \nonumber
    &= \left(-\frac{1}{2} P(\lambda)^{\dagger} q(\lambda)\right)^T P(\lambda) \left(-\frac{1}{2} P(\lambda)^{\dagger} q(\lambda)\right) \\ \nonumber 
    &+  q(\lambda)^T \left(-\frac{1}{2} P(\lambda)^{\dagger} q(\lambda)\right) + r(\lambda).
\end{align}

Finally the dual function is

\begin{align}
\inf_{\mathbf{s} \in \mathbb{R}^n} L(\mathbf{s}, \lambda)= \nonumber 
\end{align}

\begin{align}
\begin{cases} 
r(\lambda) - \frac{1}{4} q(\lambda)^T P(\lambda)^{\dagger} q(\lambda), & \text{if } P(\lambda) \succ 0 \text{ and } q(\lambda) \in \mathcal{R}(P) \\
-\infty, & \text{otherwise.}
\end{cases}
\end{align}

Based on the \textbf{lower bound property}, if $\lambda \geq 0$, we have
\begin{align}
    \mathbf{s}^* \geq r(\lambda) - \frac{1}{4} q(\lambda)^T P(\lambda)^{\dagger} q(\lambda) \nonumber
\end{align}

\textbf{Proof:} If $\tilde{\mathbf{s}}$ is feasible and $\lambda \geq 0$, then
\begin{align} 
    F(\tilde{\mathbf{s}}) \geq L(\tilde{\mathbf{s}}, \lambda) \geq \inf L(\mathbf{s}, \lambda) = g(\lambda)
\end{align}

The dual function is given by
\begin{align}
g(\lambda) &= \inf_{s \in \mathbb{R}^n} L(\mathbf{s}, \lambda) \nonumber \\
&= -\frac{1}{4} 
\left( q_0 + \sum_{i=1}^{m} \lambda_i q_i \right)^T 
\left( P_0 + \sum_{i=1}^{m} \lambda_i P_i \right)^{\dagger} \nonumber \\
&\left( q_0 + \sum_{i=1}^{m} \lambda_i q_i \right)  \quad - \sum_{i=1}^{m} \lambda_i M + h_0. 
\end{align}

We can now form the dual of the problem using Schur complements \cite{ando1979generalized}:
\begin{align}
& \nonumber \text{maximize} \quad \Gamma - \sum_{i=1}^{m} \lambda_i M + h_0 \\
& \nonumber \text{s.t.} \quad  
    \begin{bmatrix}
    P_0 + \sum_{i=1}^{N} \lambda_i P_i & \frac{1}{2} \left( q_0 + \sum_{i=1}^{N} \lambda_i q_i \right) \\
    \frac{1}{2} \left( q_0 + \sum_{i=1}^{N} \lambda_i q_i \right)^T & -\Gamma
    \end{bmatrix} \succeq 0, 
\\
    &~~~~~\quad \lambda_i \geq 0, \quad i = 1, \ldots, N, 
\end{align}

\noindent where the variable is $ \lambda \in \mathbb{R}^N $.

This turns the original nonconvex QCQP into a convex optimization program, specifically a semidefinite program (SDP). 
The SDP formulation provides a computationally efficient mechanism for obtaining a lower bound on the optimal value of the original nonconvex QCQP. In particular, the Lagrangian relaxation replaces the nonconvex quadratic constraints with linear matrix inequalities (LMIs), enabling the use of robust optimization solvers specifically designed for SDPs.

\subsection{Communication}

After selecting the vehicles for cooperative perception, we optimize the allocation of available transmission power and resource blocks (RBs) among the vehicles, accommodating imperfect communications. Our approach is to maximize channel throughput which translates to an elevated cooperative perception quality, as confirmed by our simulation results in section \ref{sec:simulations}. 

Despite evaluating the SPS algorithm, resource collisions remain a concern in V2V communication within the sidelink NR-V2X framework. This challenge arises from the stochastic nature of RB selection, where collisions occur when multiple vehicles opt for the same resource. The SPS mechanism prioritizes the lowest 20\% of candidate resources based on eceived Signal Strength Indicator (RSSI), which can cause vehicles to concentrate their selections, mainly when the channel busy ratio (CBR) is high. This leads to a higher probability of overlapping selections among vehicles in dense traffic conditions. Although the random selection process within the candidate RBs helps reduce collisions, the likelihood of interference remains notable, especially when multiple vehicles reselect resources within short time frames \cite{jeon2020explicit}.

Let $w_T$ denote the total number of available RBs within the selection window. Due to the random selection process performed by multiple vehicles, overlapping RB allocations can occur, leading to reduced communication reliability and overall network performance. More specifically, this can be expressed as:

\begin{equation}
w_T = \theta \times W_{subCh} \times (1 - \text{CBR}),
\end{equation}

\noindent where $\text{CBR}$ is Channel Busy Ratio, $\theta$ is interval between each transmission and $W_{\text{subCh}}$ is Number of subchannels on the sidelink band.

The probability of choosing a particular RB in the selection window is:
\begin{equation}
P_{\text{rb}} = \frac{1}{w_T},
\end{equation}

The probability of collision among vehicles transmitting within the same subframe can be expressed as:
\begin{equation}
\delta_{\text{COL}} = 1 - \left(1 - P_{\text{rb}}\right)^{M - 1},
\end{equation}

\noindent where \(M\) is the number of vehicles selected in the previous selection scenario.

In CP, the inter-vehicle distance is a critical factor because longer distances can increase the transmission error and it can significantly impact network performance by introducing errors. Consequently, the received signal power in C-V2X communication must be meticulously characterized \cite{gonzalez2018analytical}. The received signal power, $P_i^{\text{rx}}$, can be expressed as a function of the transmit power, $P_i^{\text{tx}}$, the path loss, $PL(d_i)$ and additional factors such as shadowing and fading. The model is given by:

\begin{align}
    P_i^{\text{rx}} = P_i^{\text{tx}} - PL(d_i),
\end{align}

\noindent where $d_i$ denotes the distance between the transmitter and receiver. The path loss is typically modeled as:
\begin{align}
PL(d_i) = 10 \log_{10}(d_i^\gamma) + L_0 + SH,
\end{align}
where $d_i$ is distance between vehicles, $\gamma$ is path loss exponent, typically a number between 2 and 4, and $SH$ is shadowing.

Shadowing $SH$ represents the effect of obstacles on signal attenuation and is modeled with a log-normal random distribution with zero mean and variance \(\sigma\). The received signal power \(P_i^{\text{rx}}\) at the receiver is hence a random variable, expressed as:
\begin{align}  
P_i^{\text{rx}} = P_i^{\text{tx}} - 10 \log_{10}(d_i^\gamma) - SH,
\end{align}

\noindent all variables are in dB.

The probability that the received signal power is lower than the sensing power threshold \(P_{\text{SEN}}\) is:
\begin{align}   
\delta_{\text{SEN}} = \int_{-\infty}^{P_{\text{SEN}}} f_{P_i^{\text{rx}}}(p) \, dp,
\end{align}

\noindent where \(f_{P_i^{\text{rx}}}(p)\) represents the PDF of the received signal power at a distance \(d_i\). Since shadowing follows a log-normal distribution, the PDF of the received signal power is given by:
\begin{align}  
f_{P_i^{\text{rx}}}(p) &= \\
&\frac{1}{\sigma \sqrt{2\pi}} \exp\left( -\left( \frac{P_i^{\text{tx}} - 10 \log_{10}(d_i^\gamma) - SH - P_{\text{SEN}}}{\sigma \sqrt{2}} \right)^2 \right), \nonumber
\end{align}

Combining the results, the probability that the received signal power at \(d_i\) is lower than the sensing power threshold is:
\begin{align} 
\delta_{\text{SEN}} &=  \\
&\frac{1}{2} \left( 1 - \operatorname{erf}\left( \frac{P_i^{\text{tx}} - 10 \log_{10}(d_i^\gamma) - SH - P_{\text{SEN}}}{\sigma \sqrt{2}} \right) \right), \nonumber
\end{align}

where \(\operatorname{erf}\) is the error function.

The total error, combining the collision error $\delta_{\text{COL}, i}$ and transmission errors $\delta_{\text{SEN}, i}$ for user $i$ is given by
\begin{equation}
\delta_{\text{Er}, i} = \delta_{\text{COL}, i} \times \delta_{\text{SEN}, i}.
\end{equation}

According to our problem, CP, and the selection scenario, these two errors can affect data transmission, and we consider both of them in the analysis. In the context of CP, both types of errors influence the reliability of data transmission and are therefore incorporated into the analysis. The throughput $\zeta_i$ for vehicle $i$ is defined based on $w_i$, representing the number of RBs allocated to a vehicle while also accounting for transmission errors characterized by the C-V2X error model:

\begin{align}
\zeta_i = R_{\text{ch}} \times w_i \times (1 - \delta_{\text{Er}, i}),
\end{align}

\noindent where $R_{\text{ch}}$ is the channel gain.

The above derivations are for given power use, and assuming equal power consumption by all vehicles. However, we can allocate the total power budget more efficiently among vehicles, for enhanced performance.

The energy use by each vehicle can be modeled as

\begin{equation}
\mathbb{E}_i = \frac{P_i^{\text{tx}} T}{1 - \delta_{\text{Er}, i}},
\end{equation}
\noindent where $T$ is the transmission interval, and the $1 - \delta_{\text{Er}, i}$ term is to compensate for sending extra bits (replacing erroneous bits).

Our goal is to maximize the ratio of throughput to energy consumption:
\begin{equation}
\sum_{i=1}^{M} \frac{\zeta_i}{\mathbb{E}_i},
\end{equation}
by adjusting the number of radio blocks and power share for each user.

\noindent This can be rewritten so that we drop the constant $T$ for brevity: 
\begin{align}
    \text{maximize} \quad \sum_{i=1}^{M} \frac{R_{\text{ch}} \times w_i \times (1 - \delta_{\text{Er}, i})}{\frac{P_i^{\text{tx}}}{1 - \delta_{\text{Er}, i}}},
\end{align}

\noindent or more compactly:
\begin{align}
    \text{maximize} \quad \sum_{i=1}^{M} \frac{R_{\text{ch}} \times w_i \times (1 - \delta_{\text{Er}, i})^2}{P_i^{\text{tx}}}.
\end{align}

Then we plug collision error and transmission error

\begin{align} 
 \text{maximize} \quad \sum_{i=1}^{M} & \Bigg(R_{\text{ch}} \times w_i \times \Big( 1 - (\delta_{\text{COL}, i} \times \delta_{\text{SEN}, i}) \Big)^2 \nonumber \\
& \Bigg/  P_i^{\text{tx}} \Bigg). 
\end{align}

Our goal is to maximize the following objective function:

\begin{align}
\max &\Bigg[ \sum_{i=1}^{M} \Bigg(R_{\text{ch}} \times w_i \times 
\Bigg( (1 - P_{\text{res}})^{M - 1} \\ \nonumber
&- \frac{1}{2} \left( 1 - \operatorname{erf} \left( 
\frac{P_i^{\text{tx}} - 10 \log_{10}(d_i^\gamma) - SH - P_{\text{SEN}}}{\sigma \sqrt{2}} 
\right) \right) \Bigg)^2  \\ \nonumber
& \Bigg/ P_i^{\text{tx}} \Bigg)\Bigg], 
\end{align}

\noindent subject to the following constraints:
\begin{align} 
\sum_{i=1}^{M} P_i^{\text{tx}} &\leq P_T, \\
\sum_{i=1}^{M} w_i &= w_T,
\end{align}

\noindent where $P_T$ is the total power transmission, and $w_i$ is the number of radio blocks for user $i$.

The constraints are both affine and convex. To simplify the error function, we apply a Taylor series expansion to the error function:
\begin{align} 
    \operatorname{erf}(\mathbb{Q}) = \frac{2}{\sqrt{\pi}} \int_0^\mathbb{Q} e^{-\mathcal{B}^2} \, d\mathcal{B}.
\end{align}

Using 
\begin{align} 
    e^{-\mathcal{B}^2} = 1 - \mathcal{B}^2 + \frac{\mathcal{B}^4}{2!} - \frac{\mathcal{B}^6}{3!} + \cdots \quad (\mathbb{R} = \infty).
\end{align}
we can find a power series expansion for the error function, as follows
\begin{align} 
   \operatorname{erf}(\mathbb{Q}) = \frac{2}{\sqrt{\pi}} \int_0^\mathbb{Q} e^{-\mathcal{B}^2} \, d\mathcal{B} 
   = \frac{2}{\sqrt{\pi}} \sum_{n=0}^{\infty} \frac{(-1)^n \kappa^{2n+1}}{n! (2n + 1)}.
\end{align}

\noindent Consequently, the objective function becomes 

\begin{align}
\sum_{i=1}^{M} & \Bigg(R_{\text{ch}} \times w_i \times 
\Big[ (1 - P_{\text{res}})^{M - 1} \nonumber \\
&- \frac{1}{2} \left( 1 - \frac{2}{\sqrt{\pi}}\left( 
\frac{P_i^{\text{tx}} - 10 \log_{10}(d_i^\gamma) - SH - P_{\text{SEN}}}{\sigma \sqrt{2}} 
\right) \right) \Big ]^2 \nonumber \\
& \Big/ P_i^{\text{tx}}\Bigg).
\end{align}

The optimization problem is fractional, making it inherently non-convex. To address this matter, we employ Dinkelbach's algorithm, which is well-suited for handling fractional programming problems. 
It iteratively approximates the original objective function with a linear model and determines the search direction by solving a linearized optimization problem over the feasible domain. Thereby, it 
turns the problem into an iterative optimization problem with convex objective and constraint functions. 
Subsequently, we use Frank–Wolfe's algorithm \cite{frank1956algorithm}, also known as the conditional gradient method, to solve these subproblems. Frank–Wolfe's algorithm is known for its high efficiency in solving constrained convex optimization problems, particularly when direct projection onto the constraint set is computationally expensive or challenging. This step involves finding a descent direction using the gradient of the function, followed by determining an appropriate step size, and updating the solution iteratively.

In essence, our algorithm combines Dinkelbach and Frank–Wolfe algorithms into a unified framework where the outer and inner loops, respectively, utilize Dinkelbach and Frank–Wolfe algorithms to efficiently determine the optimal solution.
The summary of our algorithm is given in Algorithm 1.
This novel combination not only guarantees convergence to the optimal solution but also significantly enhances the tractability of resource allocation in wireless networks.

\begin{algorithm}
\caption{Combined Dinkelbach and Frank–Wolfe Algorithm}
\label{alg:dinkelbach-frank-wolfe}
\begin{algorithmic}[1]
\Require Fractional objective \( \frac{\mathcal{N}(\chi)}{\mathcal{D}(\chi)} \), initial point \( \chi_1 \in \mathbb{S} \), tolerance \( \epsilon > 0 \)
\State Set \( \eta_2 = \frac{\mathcal{N}(\chi_1)}{\mathcal{D}(\chi_1)} \) (or \( \eta_2 = 0 \)), and initialize \( k = 2 \).

\While{True}
    \State Solve the Subproblem using Frank–Wolfe Algorithm
    \State Maximize the linearized objective \( \mathcal{F}(\eta_k) = \mathcal{N}(\chi) - \eta_k \mathcal{D}(\chi) \) over the feasible set \( \mathbb{D} \) using the Frank–Wolfe method:
    
    \State Find \( \psi_j \in \mathbb{S} \) such that \( \psi_j = \arg \min_{\psi \in \mathbb{S}} \psi^T \nabla f(\chi_j) \)
    \State Determine step size \( m_j = \frac{2}{j + 2} \) 
    \State Update the solution:
        \[
        \chi_{j+1} = \chi_j + m_j (\psi_j - \chi_j)
        \]
    \State Increment \( j \leftarrow j + 1 \) until convergence within tolerance.

    \State Denote the optimal solution as \( \chi_k \).

    \State Check Optimality Condition for Dinkelbach
    \If{ \( \mathcal{F}(\eta_k) < \epsilon \) }
        \State Output \( \chi_k \) as the optimal solution and stop.
    \Else
        \State Update the parameter:
        \[
        \eta_{k+1} = \frac{\mathcal{N}(\chi_k)}{\mathcal{D}(\chi_k)}
        \]
        \State Set \( k \leftarrow k + 1 \) and repeat from Step 1.
    \EndIf
\EndWhile

\Return Optimal solution \( \chi^* = \chi_k \)
\end{algorithmic}
\end{algorithm}

\subsection{Fusion}

Our implementation facilitates various fusion approaches to form an optimal cooperative perception with candidate vehicles, under constrained resources. In this work, we adopt a late fusion approach, primarily for ease of implementation and improved robustness. Each vehicle employs YOLOv8 for object detection, and the ego vehicle subsequently gathers object detection results from helper vehicles. To enhance the visual range and improve detection accuracy, the ego vehicle applies a fusion strategy, such as a majority voting rule, to integrate the helper detections effectively. For the fusion part, we select the best IoU between the ego vehicle and the helper, to investigate the optimality of the selected helper. 
 
\vspace{-0.1cm}
\begin{align}
    IoU_s = Max(IoU_e, IoU_h)
\end{align}

\begin{figure*}[!htbp]
    \centering
    \includegraphics[width=\textwidth, height=0.8\textheight, keepaspectratio]{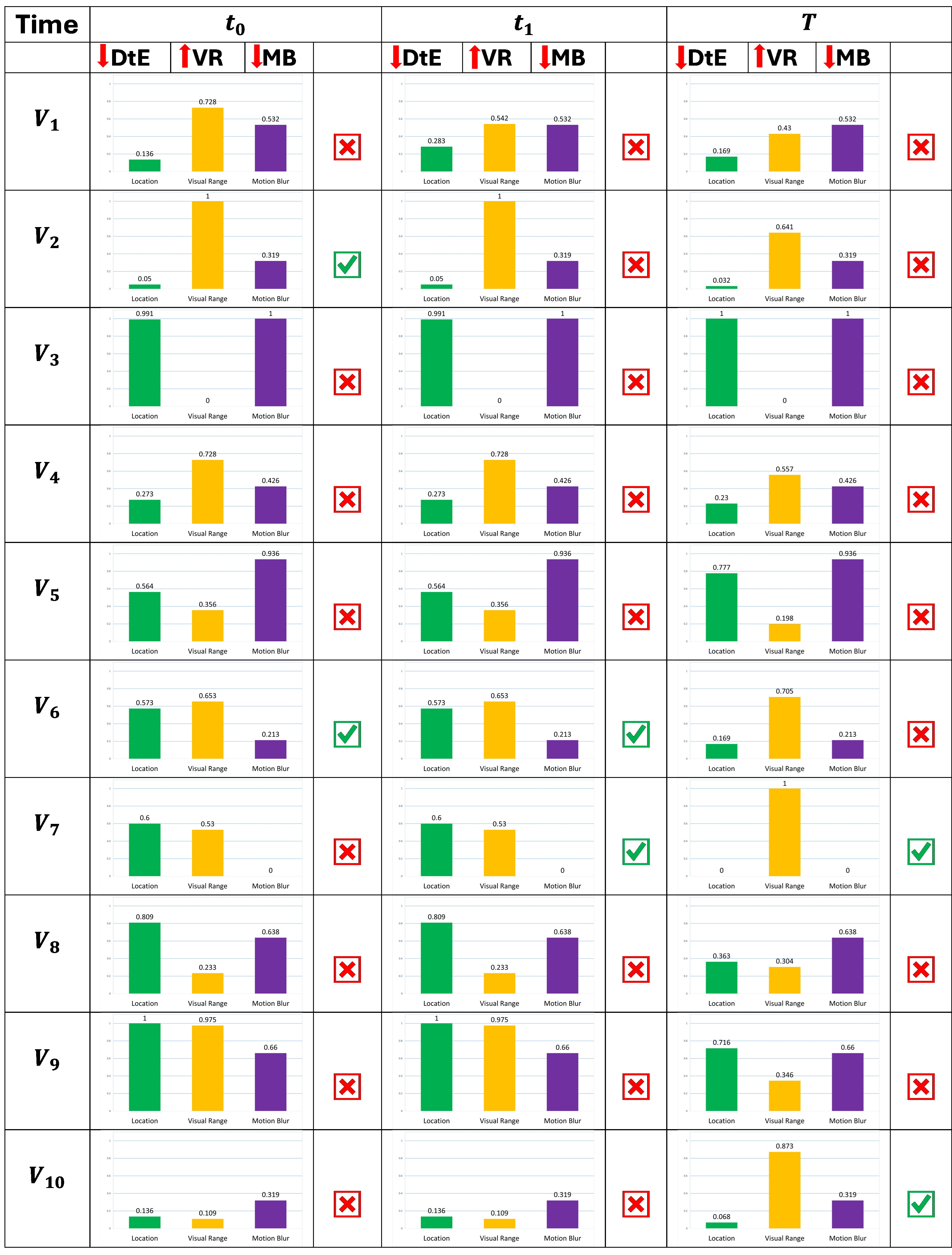}
    \caption{Optimization of helper vehicle selection over time intervals in an exemplary scenario, according to distance to ego vehicle (DtE), visual range (VR), and motion blur (MB). At $t_0$, the vehicles $V_2$
and $V_6$ are the optimal helpers, whereas at $t_1$, $V_6$ and $V_7$
 become the best choices due to changes in vehicle locations and visual ranges. Selecting helper vehicles dynamically at fixed intervals $T$ ensures robustness, as demonstrated by $V_7$ and $V_{10}$ being the optimal choices over the interval.}
    \label{fig:tb1}
\end{figure*}

\section{Simulation}   \label{sec:simulations}
In this section, we evaluate the performance of the proposed cooperative perception method in comparison with a single-vehicle perception system as well as cooperative systems with random and uniform selection. More specifically, we inspect the selected vehicles in terms of their collective visual range, motion blur, and ultimate object detection accuracy.  Furthermore, we investigate the optimality of throughput over energy consumption, under imperfect communication conditions. The experimental setup considers $N=10$ vehicles, where $M$ vehicles are selected for optimization. The spatial distribution of the vehicles is modeled using the Point Poisson Process, while their velocities follow a truncated Gaussian distribution, as declared in section \ref{sec:system_model}. 
The optimization is performed over a time interval $T$, during which 
the visual range and the distances between helper vehicles and the ego vehicle are subject to substantial change. For simplicity, we assume that vehicle velocities remain constant within the interval $T$.
In the initial evaluation phase, we compare the proposed approach with alternative vehicle selection methods to evaluate the efficacy of our proposed helper selection strategy. Quantitative metrics are employed to assess the improvements in visual range and the reduction in motion blur, alongside higher throughput and energy efficiency.

\begin{figure}[h]
\centering
\begin{subfigure}[b]{0.48\textwidth}
\centering
\includegraphics[width=1\columnwidth]{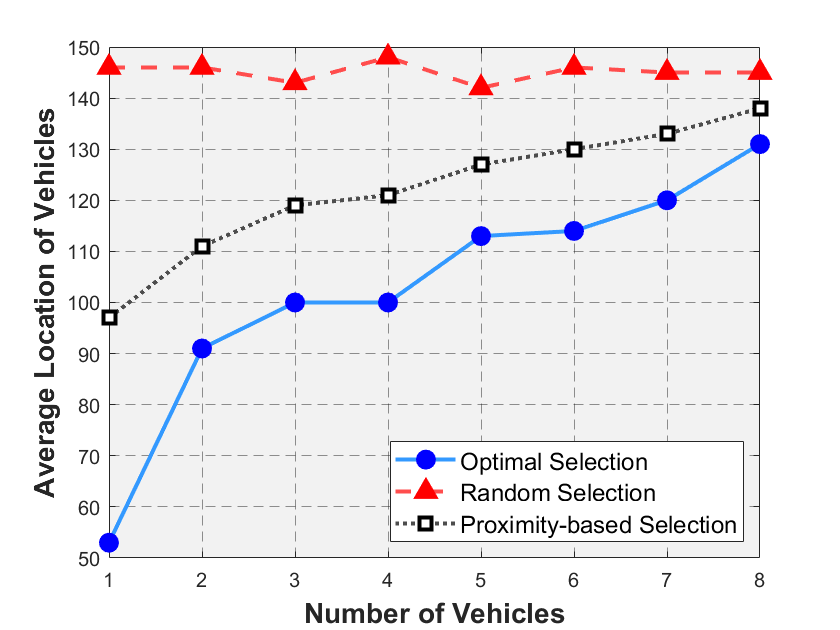}
\caption{}
\end{subfigure}
\begin{subfigure}[b]{0.48\textwidth}
\centering
\includegraphics[width=1\columnwidth]{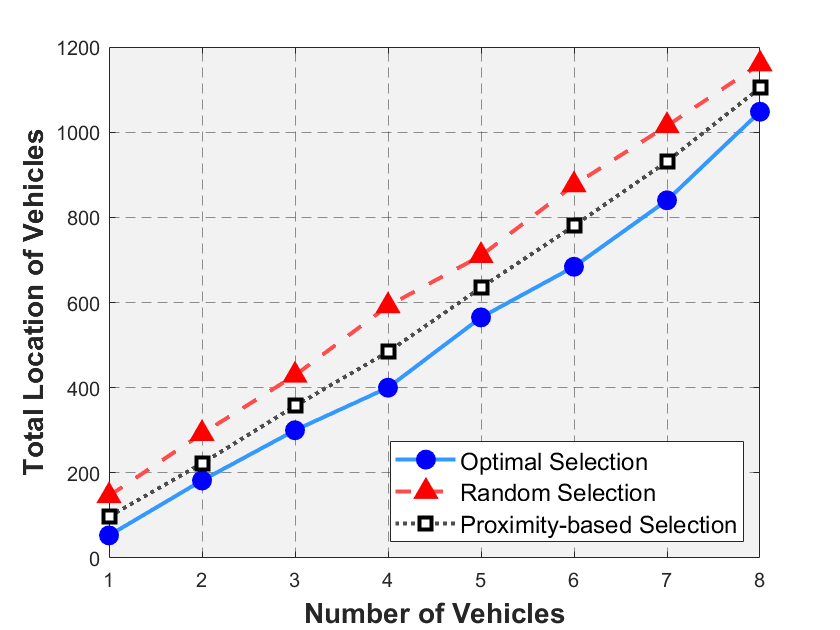}
\caption{}
\end{subfigure}
\caption{Comparison of helper vehicle selection methods based on location metrics (a) Average location (distance to the ego vehicle) versus the number of selected helper vehicles. (b) Total location (cumulative distance to the ego vehicle) versus the number of selected helper vehicles. The results demonstrate that our proposed method consistently selects helper vehicles closer to the ego vehicle compared to random selection and proximity-based methods.}
\label{fig:location}
\end{figure} 

Fig.\ref{fig:tb1} 
presents two facts simultaneously. The first observation highlights that our proposed method evaluates and selects the optimal helper vehicle by considering multiple factors. For instance, suppose we are to select two helper vehicles. 
For instance, at time $t_0$, vehicles $V_7$ and $V_9$, respectively, exhibit the best performance in terms of lower motion blur, prolonged visual range, and lower distance to ego vehicle. However, based on an overall evaluation that balances these factors, vehicles $V_2$ and $V_6$ emerge as the first two top choices.

The second fact is about the rationale behind optimizing helper vehicle selection within an interval of $T$, to account for the dynamicity of the situation which undermines the optimality of epoch-based optimization. For instance, recall the same problem of selecting two best helpers. At time $t_0$, vehicles $V_2$ and $V_6$ serve as optimal helpers, while at $t_1$, the best choices shift to $V_6$ and $V_7$. This variability underscores the dynamic nature of helper selection as vehicles' positions shift. However, selecting helper vehicles by averaging the objective function over the entire interval $T$ provides a robust approach, as demonstrated by $V_7$ and $V_{10}$ being the optimal helper across the interval. This result highlights the limitations of momentary selection and emphasizes the importance of averaging in equations (\ref{eq:f1}), (\ref{eq:f2}), (\ref{eq:f3}) to account for temporal variations in vehicle locations and ranges.

\begin{figure}[h]
\centering
\begin{subfigure}[b]{0.48\textwidth}
\centering
\includegraphics[width=1\columnwidth]{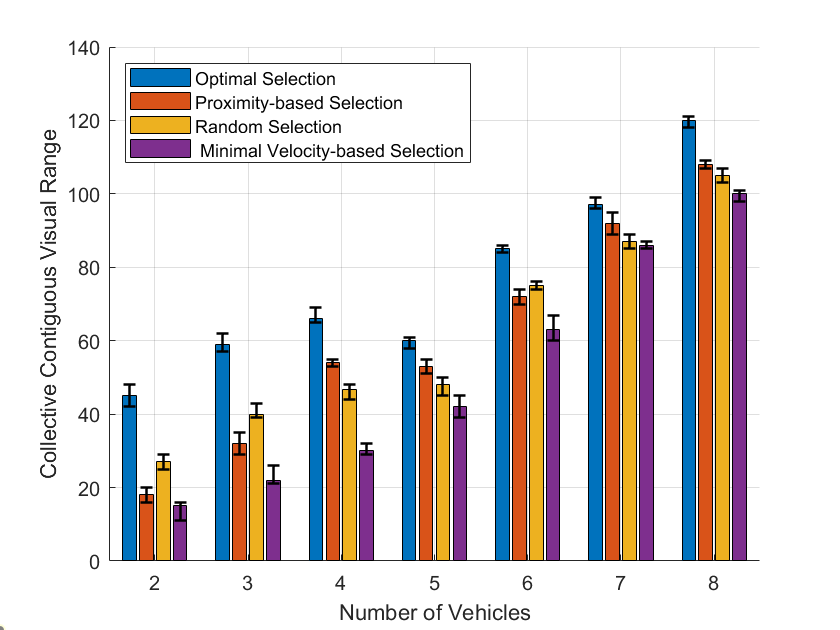}
\caption{}
\end{subfigure}
\hfill
\begin{subfigure}[b]{0.48\textwidth}
\centering
\includegraphics[width=1\columnwidth]{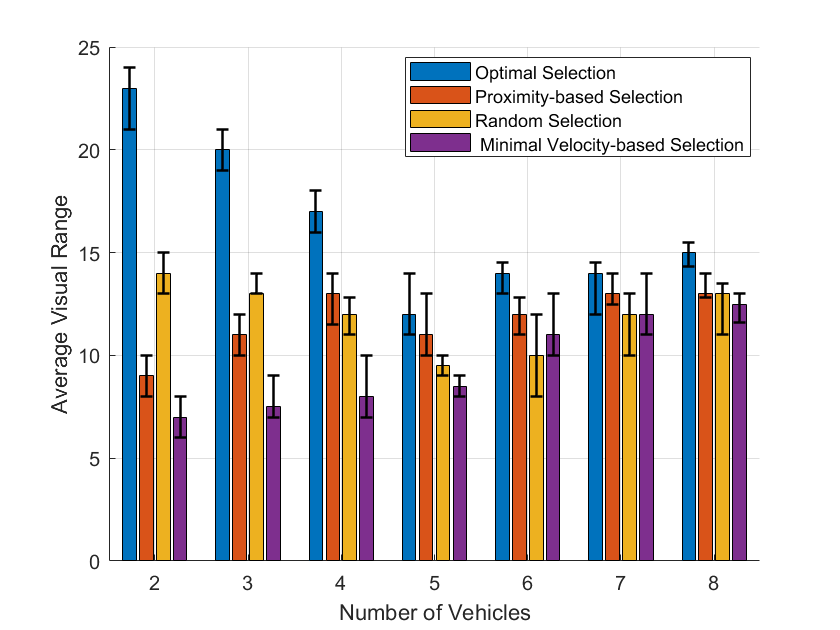}
\caption{}
\end{subfigure}
\caption{(a) The collective contiguous visual range achieved for different vehicle selection techniques. (b) The average visual range was obtained across various vehicle selection methods. The proposed method effectively identifies and recruits optimal cooperative helpers to enhance the visual range of the ego vehicle through CP mechanisms.}
\label{fig:VI-R}
\end{figure}

Fig. \ref{fig:location} represents the location of selected helper vehicles compared to random selection (with uniform distribution) and proximity-based selection (selecting closer vehicles to the ego vehicle at the beginning of the epoch methods. The location metric represents the distance to the ego vehicle, with shorter distances being more desirable. The subfigures (a) and (b) illustrate the comparative performance of the methods. Our proposed method selects optimal helpers over an interval $T$, resulting in helper vehicles consistently closer to the ego vehicle, as indicated by the lower distance values. This demonstrates the effectiveness of our method in selecting the most suitable helper vehicles for interval-based optimization.

\begin{figure}[htbp]
\begin{center}
\centerline{
\includegraphics[width=0.6\columnwidth]{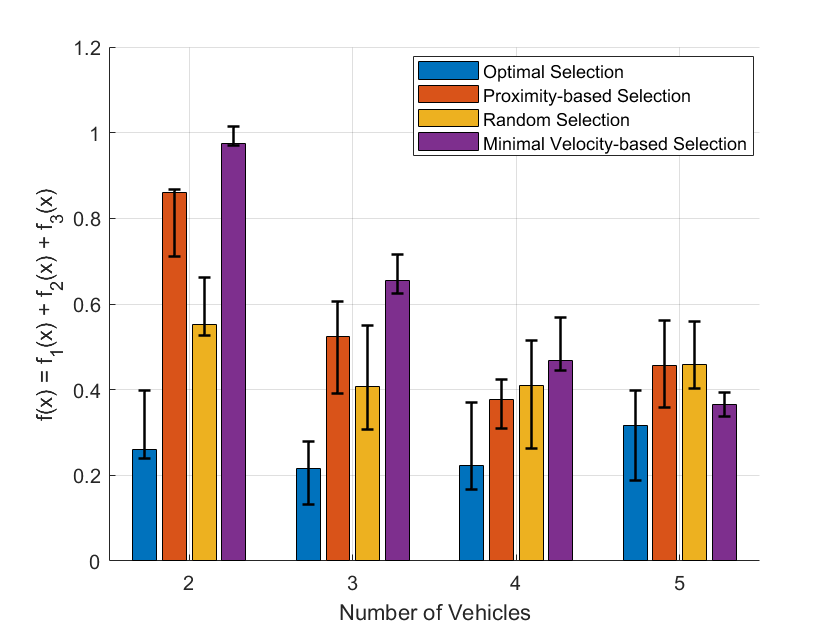}}
\vspace{0.2cm}
\caption{Comparison of the objective function for helper vehicles selected by the proposed method and other selection methods. The proposed method achieves significantly better selection, demonstrating its superiority.}
\label{fig:R4}
\end{center}
\end{figure} 

Total visual range achieved through different helper vehicle selection methods, as represented in Fig. \ref{fig:VI-R}. The proposed approach selects optimal helpers in terms of higher collective visual range, consistently outperforming proximity-based selection, random selection, and minimal velocity-based selection (vehicles with lower velocity at the beginning of the epoch). By optimizing over the interval $T$, the method ensures superior coverage and highlights its effectiveness compared to alternative methods.

\begin{figure}[htbp]
\begin{center}
\centerline{
\includegraphics[width=0.6\columnwidth]{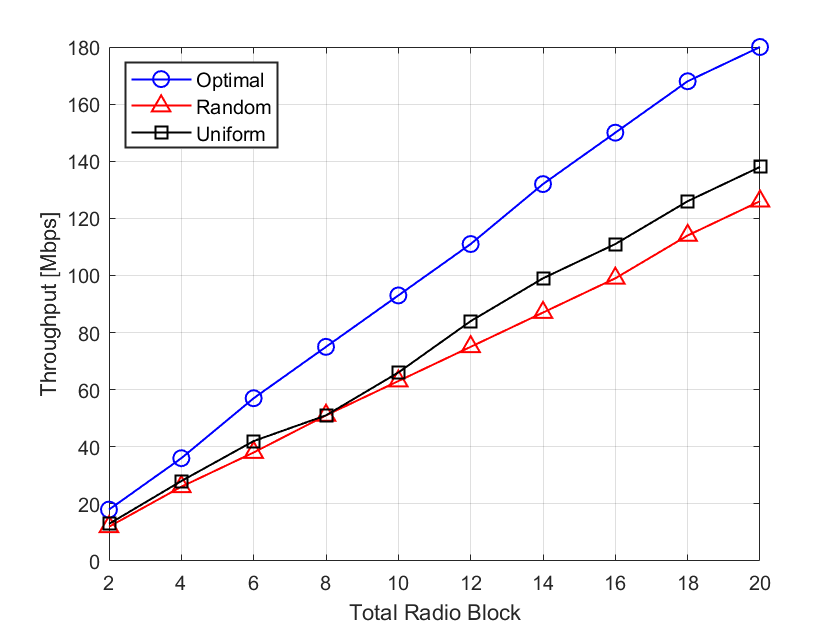}}
\vspace{0.2cm}
\caption{Throughput as a function of total RBs: The proposed algorithm consistently achieves higher throughput than random and uniform RB allocation strategies.}
\label{fig:T1}
\end{center}
\end{figure} 

\begin{figure}[htbp]
\begin{center}
\centerline{
\includegraphics[width=0.6\columnwidth]{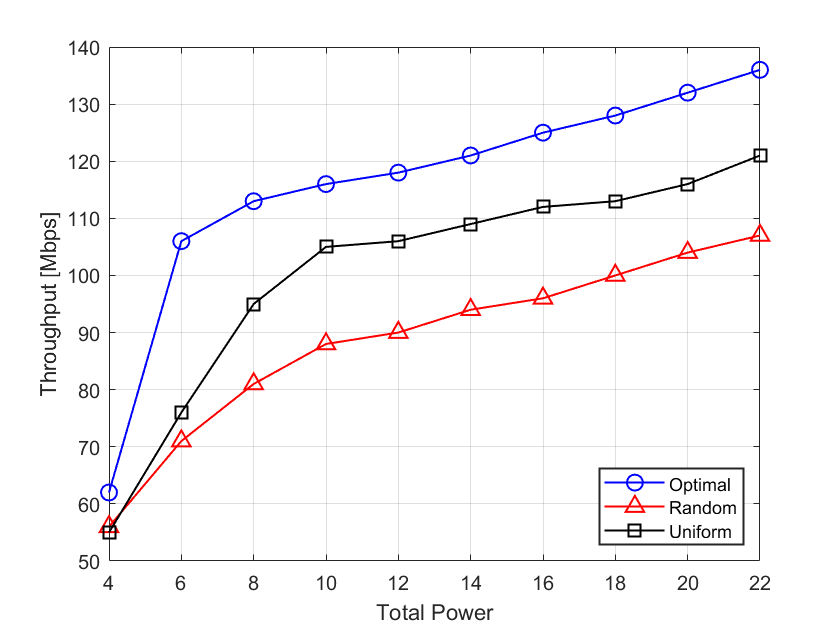}}
\vspace{0.2cm}
\caption{Throughput as a function of total RBs. The proposed algorithm consistently reaches higher throughput than random and uniform RB allocation techniques.}
\label{fig:T2}
\end{center}
\end{figure} 

Fig. \ref{fig:R4} illustrates the objective function $f(x)$ in equation (\ref{eq:optimization}), which was the ground for selecting the best helpers. The results in this figure confirm that our method consistently selects the best helper, as was intended. It also shows that recruiting more helpers can enhance the overall quality of perception if one can afford the incurred complexity and communication overhead. However, this improvement saturates, after selecting the first two or three helper vehicles. 

These findings highlight the effectiveness of the proposed method in enhancing visual clarity and enabling reliable perception in dynamic environments. 

\begin{figure}[h]
\centering
\begin{subfigure}[b]{0.5\textwidth}
\centering
\includegraphics[width=1\columnwidth]{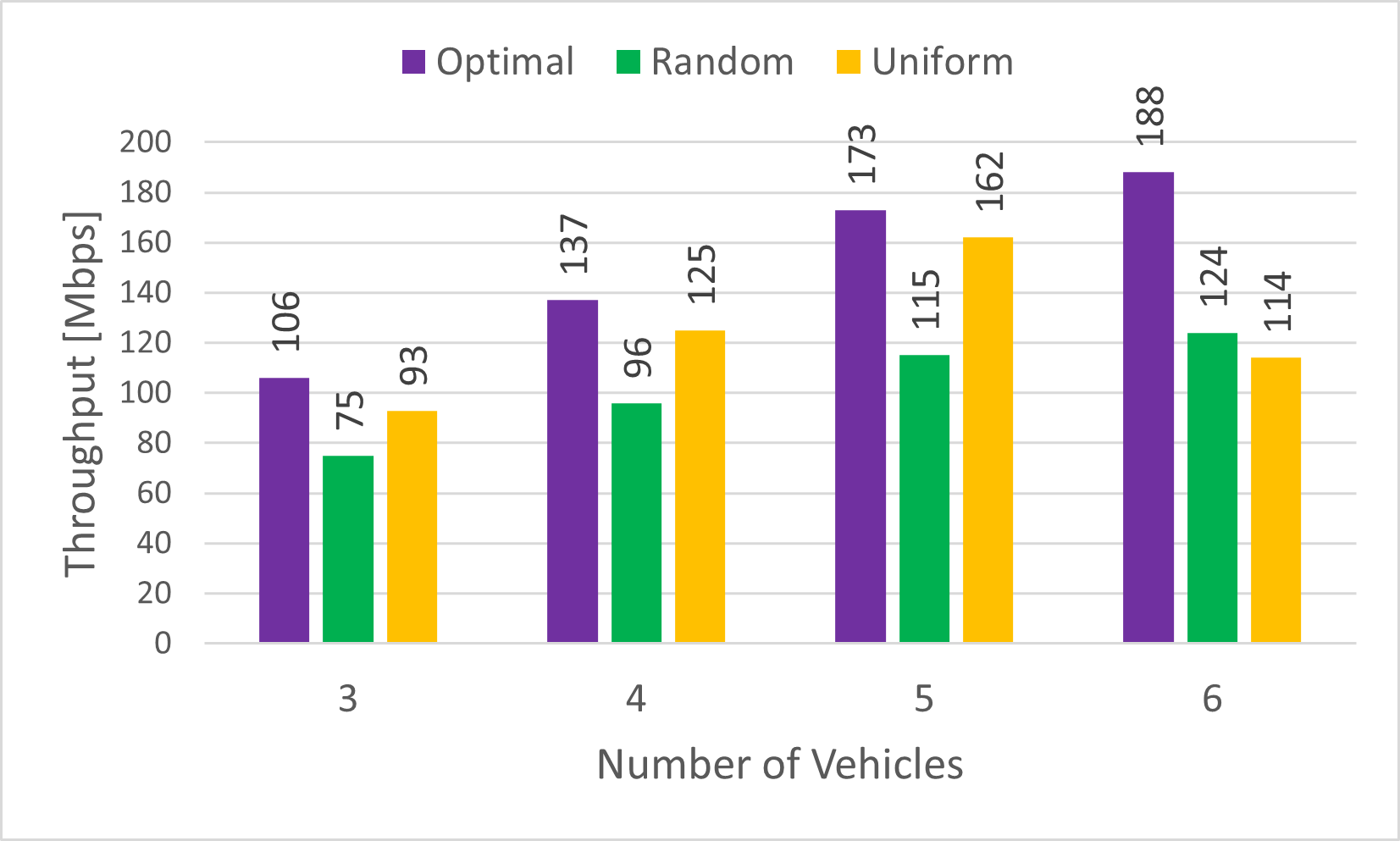}
\caption{}
\end{subfigure}
\hfill
\begin{subfigure}[b]{0.5\textwidth}
\centering
\includegraphics[width=1\columnwidth]{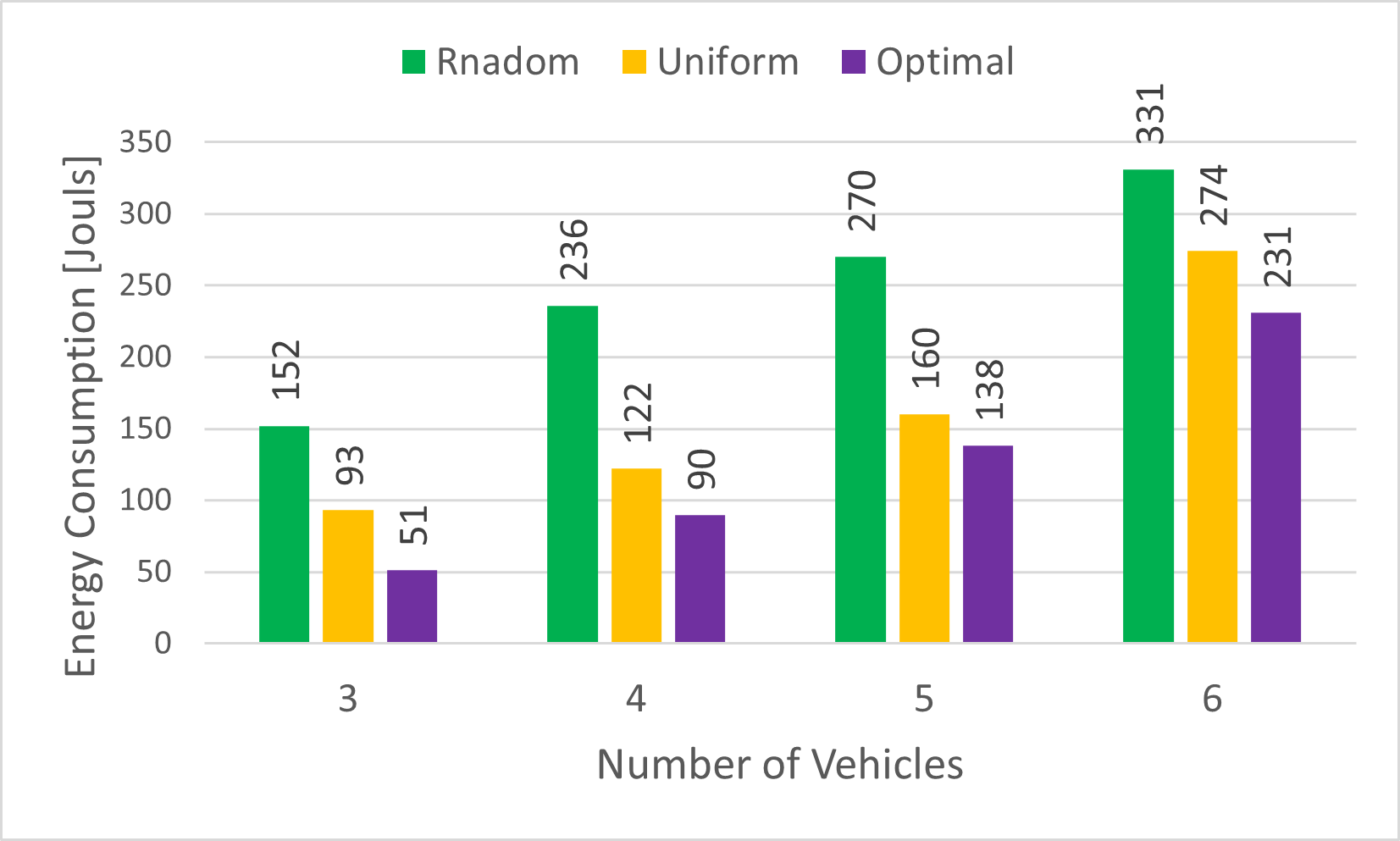}
\caption{}
\end{subfigure}
\caption{(a) Throughput (b) Energy consumption as a function of the number of vehicles. The proposed algorithm demonstrates superior performance across varying vehicles compared to alternative methods.}
\label{fig:T3}
\end{figure} 

We further investigate the second part of our framework, which aims to 
optimize throughput and energy consumption by 
efficiently allocating resources and transmission power among vehicles.
Fig.\ref{fig:T1} compares the performance of the proposed algorithm with random and uniform RB assignment strategies. The results demonstrate that the proposed method achieves significantly higher channel throughput than random and uniform RB allocation, highlighting its effectiveness in resource management.

Fig. \ref{fig:T2} illustrates the channel throughput as a function of total transmission power allocated to vehicles. The proposed algorithm consistently outperforms random and uniform allocation methods, demonstrating outstanding performance.
Figs. \ref{fig:T2} and \ref{fig:T3} collectively highlight the ability of the proposed algorithm to maximize channel throughput, a critical factor in CP. Even with the selection of optimal helper vehicles, imperfect communication can hinder the transmission of high-quality images, making CP ineffective. These results emphasize the importance of efficient resource allocation to ensure reliable and robust CP.

Fig.\ref{fig:T3} (a) expresses the channel throughput as a function of the number of vehicles. The results demonstrate that the proposed algorithm maintains higher throughput across varying numbers of vehicles than random and uniform RB allocation methods by optimally assigning RBs among all vehicles.
Fig. \ref{fig:T3} (b) shows the energy consumption of the channel as a function of the number of selected vehicles. The proposed algorithm consistently consumes less energy than random and uniform methods, demonstrating its efficiency.

The above simulations confirm that the proposed method is practical, and further enhances the quality of formed CP by maximizing its energy efficiency via optimal transmission power and RB allocation. This approach achieves higher throughput and significantly lowers energy consumption, making it a highly impactful solution for efficient and sustainable vehicle communication networks.

\begin{figure}[htbp]
    \centering
    \includegraphics[width=\columnwidth]{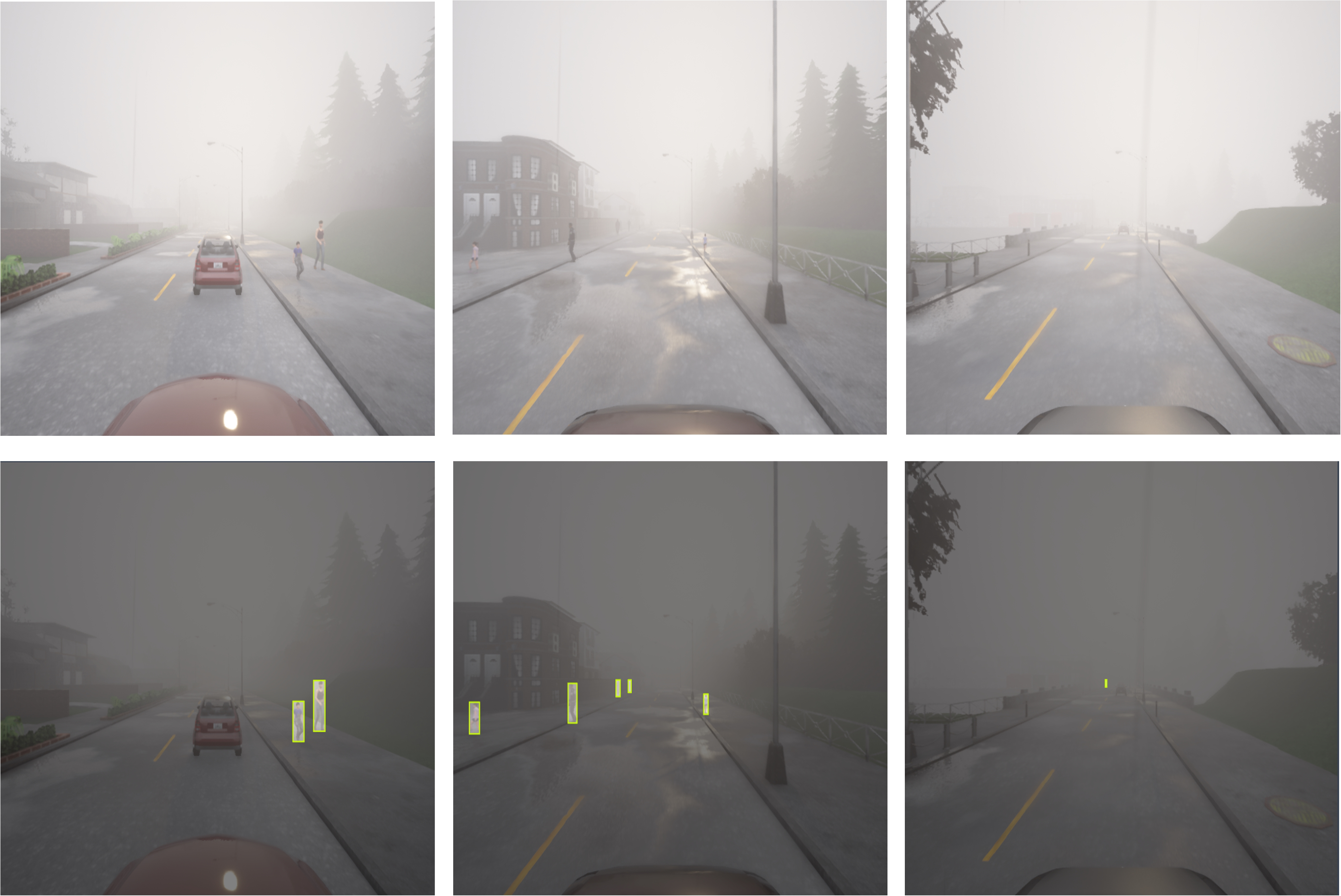}
    \caption{Some examples of our generated dataset.}
    \label{fig:Dataset}
\end{figure}

Finally, in order to inspect the applicability of the proposed system in more realistic scenarios, we perform object detection via cooperative perception as an exemplary application. To this end, 
we trained our model using scenarios generated by the CARLA simulator. Since no publicly available dataset exists for visual CP, we developed our own dataset which comprises over 2,000 synthetic images. The scenarios include one ego vehicle and three helper vehicles in foggy weather conditions. The dataset was manually annotated, and the YOLOv8 model was employed to detect pedestrians under these challenging conditions. 
Table \ref{tab:od_metrics1} presents the accuracy of OD under perfect communication conditions, comparing scenarios where the ego vehicle recruits helper vehicles versus scenarios where each vehicle (ego and helpers) individually detects the target object. The results highlight that recruiting helpers significantly improves OD accuracy. Specifically, Helper 2 achieves the best results, while Helper 3 underperforms in terms of IoU, recall, and F1 score due to its greater distance from the ego vehicle.
We introduce a packet drop rate to simulate imperfect communication and analyze its impact on OD accuracy. Table \ref{tab:od_metrics2} summarizes these results, demonstrating that the proposed method maintains the highest accuracy despite imperfect communication conditions. These results demonstrate the robustness of the proposed method under intermittent communications scenarios.

\begin{table}[ht]
    \centering
    \begin{tabular}{c|ccccccc}
        \hline
        & EGO & H1 & H2 & H3 & EGO+H1 & EGO+H2 & EGO+H3 \\ 
        \hline
        IOU  & 0.43 & 0.68 & 0.75 & 0.43 & 0.77 & \textbf{0.88} & 0.66 \\
        RECALL  & 0.23 & 0.34 & 0.38 & 0.21 & 0.39 & \textbf{0.45} & 0.31 \\
        F1 SCORE & 0.37 & 0.5 & 0.54 & 0.35 & 0.55 & \textbf{0.61} & 0.47 \\
        \hline
    \end{tabular}
    \caption{Object detection metrics for the ego vehicle and helpers individually, compared to metrics when the ego vehicle recruits helpers.}
    \label{tab:od_metrics1}
\end{table}

\vspace{0.5cm} 

\begin{table}[ht]
    \centering
    \begin{tabular}{c|ccc}
        \hline
        & \makecell{\textbf{Proposed} \\ \textbf{Method}} 
        & \makecell{\textbf{Random} \\ \textbf{Selection}} 
        & \makecell{\textbf{Proximity-Based} \\ \textbf{Selection}} \\
        \hline
        IOU    & \textbf{0.84} & 0.54 & 0.32 \\
        RECALL & \textbf{0.42} & 0.26 & 0.19 \\
        F1 SCORE & \textbf{0.58} & 0.4 & 0.27 \\
        \hline
    \end{tabular}
    \caption{Comparison of the proposed method with alternative selection strategies under channel error conditions applied to the images.}
    \label{tab:od_metrics2}
\end{table}

\section{Conclusions} \label{sec:conclusions}

This work introduced a novel CP framework designed to enhance the perception quality of autonomous vehicles in complex and adverse scenarios. The core idea was recruiting front vehicles as helpers and executing a collective analysis of imagery to enhance situational awareness while accounting for factors like motion blur, visual range, and distance to the ego vehicle. We proposed a lightweight iterative algorithm that solves this optimization problem. This approach can be particularly advantageous in adverse weather conditions, and low vision range in curly roads and crowded traffic. In addition to proposing a framework for optimized vehicle selection, we added an additional layer of network optimization to enhance the quality of cooperative perception by optimizing throughput per unit energy consumption, under dynamic and heterogeneous networking conditions. 
We conducted extensive experiments using synthetic data generated by the CARLA simulator, demonstrating a remarkable 10\% gain in detection accuracy using our method. Key contributions include a dynamic helper selection mechanism that accounts for spatial positions, relative velocities, and channel qualities to maximize visual range and perception quality, an optimization strategy used to allocate communication resources efficiently under non-ideal conditions, and a comprehensive evaluation framework that considers real-world constraints such as packet drop rates and noisy channel conditions in modern vehicular communications, including LTE and NR-5G. This work underscores the transformative potential of CP in overcoming the limitations of individual AV perception systems, paving the way for safer and more reliable autonomous driving. Future research will explore extensions of this framework to incorporate real-time adaptation to rapid environmental changes, integration with heterogeneous sensor modalities, and large-scale field testing in real-world vehicular networks.

  \bibliographystyle{elsarticle-num} 
  \bibliography{cas-refs}

\end{document}